\pdfoutput=1
\documentclass[11pt]{article}

\usepackage{ACL2023}

\usepackage{times}
\usepackage{latexsym}

\usepackage[T1]{fontenc}
\usepackage[utf8]{inputenc}

\usepackage{microtype}

\usepackage{inconsolata}
\usepackage{xcolor, colortbl}

\newcommand{\holdout}{\cellcolor[rgb]{0.9, 0.95, 1}}
\newcommand{\hatecheck}{\cellcolor[rgb]{0.9, 1, 0.9}}
\newcommand{\hateday}{\cellcolor[rgb]{0.95, 0.9, 1}}

\definecolor{tomato}{HTML}{FF6347}
\definecolor{gold}{HTML}{FFD700}
\definecolor{darkgrey}{HTML}{A9A9A9}
\definecolor{holdout}{rgb}{0.9, 0.95, 1}
\definecolor{hatecheck}{rgb}{0.9, 1, 0.9}
\definecolor{hateday}{rgb}{0.95, 0.9, 1}

\usepackage{graphicx}
\usepackage{multirow}
\usepackage{booktabs}
\usepackage{colortbl}
\usepackage{graphicx}
\usepackage{tabularx}

\title{\textsc{HateDay}: Insights from a Global Hate Speech Dataset \\ Representative of a Day on Twitter}

\author{Manuel Tonneau\textsuperscript{\rm 1, \rm 2, \rm 3}, 
        Diyi Liu\textsuperscript{\rm 1},
        Niyati Malhotra\textsuperscript{\rm 2}, 
        Scott A. Hale\textsuperscript{\rm 1,\rm 4}, \\
        {\bf Samuel P. Fraiberger}\textsuperscript{\rm 2, \rm 3}, 
        {\bf Victor Orozco-Olvera}\textsuperscript{\rm 2},
        {\bf Paul Röttger}\textsuperscript{\rm 5}\\
        \textsuperscript{\rm 1}University of Oxford,
        \textsuperscript{\rm 2}World Bank,
        \textsuperscript{\rm 3}New York University, \\
        \textsuperscript{\rm 4}Meedan,
        \textsuperscript{\rm 5}Bocconi University
}

\begin{document}
\maketitle
\begin{abstract}

To address the global challenge of online hate speech, prior research has developed detection models to flag such content on social media.\ However, due to systematic biases in evaluation datasets, the real-world effectiveness of these models remains unclear, particularly across geographies. We introduce \textsc{HateDay}, the first global hate speech dataset representative of social media settings, constructed from a random sample of all tweets posted on September 21, 2022 and covering eight languages and four English-speaking countries.\ Using \textsc{HateDay}, we uncover substantial variation in the prevalence and composition of hate speech across languages and regions.\ We show that evaluations on academic datasets greatly overestimate real-world detection performance, which we find is very low, especially for non-European languages. Our analysis identifies key drivers of this gap, including models’ difficulty to distinguish hate from offensive speech and a mismatch between the target groups emphasized in academic datasets and those most frequently targeted in real-world settings. We argue that poor model performance makes public models ill-suited for automatic hate speech moderation and find that high moderation rates are only achievable with substantial human oversight. Our results underscore the need to evaluate detection systems on data that reflects the complexity and diversity of real-world social media.

\noindent \textcolor{red}{\textbf{Content warning:} This article contains illustrative examples of hateful content.}

\end{abstract}

%%%%%%%%%%%%%%%%%%%%%%%%%%%%%%%%%%%%%%%%%%%%%%%%%%%%%%%%%%%%%%%%%%%%%%%%%%%%%%%%%%%%%%
%%%%%%%%%%%%%%%%%%%%%%%%%%%%%%%%%%%%%%%%%%%%%%%%%%%%%%%%%%%%%%%%%%%%%%%%%%%%%%%%%%%%%%
\section{Introduction}
\vspace{-1mm}
Social media users frequently encounter hate speech in their feeds \cite{unesco2023survey}, raising concerns about its potential to incite offline violence \cite{muller2021fanning}. In response, a substantial body of research has focused on developing automated hate speech detection models \cite{vidgen2020directions}. However, prior work in this area suffers from three major limitations.  

\begin{figure}[t]
    \raggedleft
    \includegraphics[width=\columnwidth]{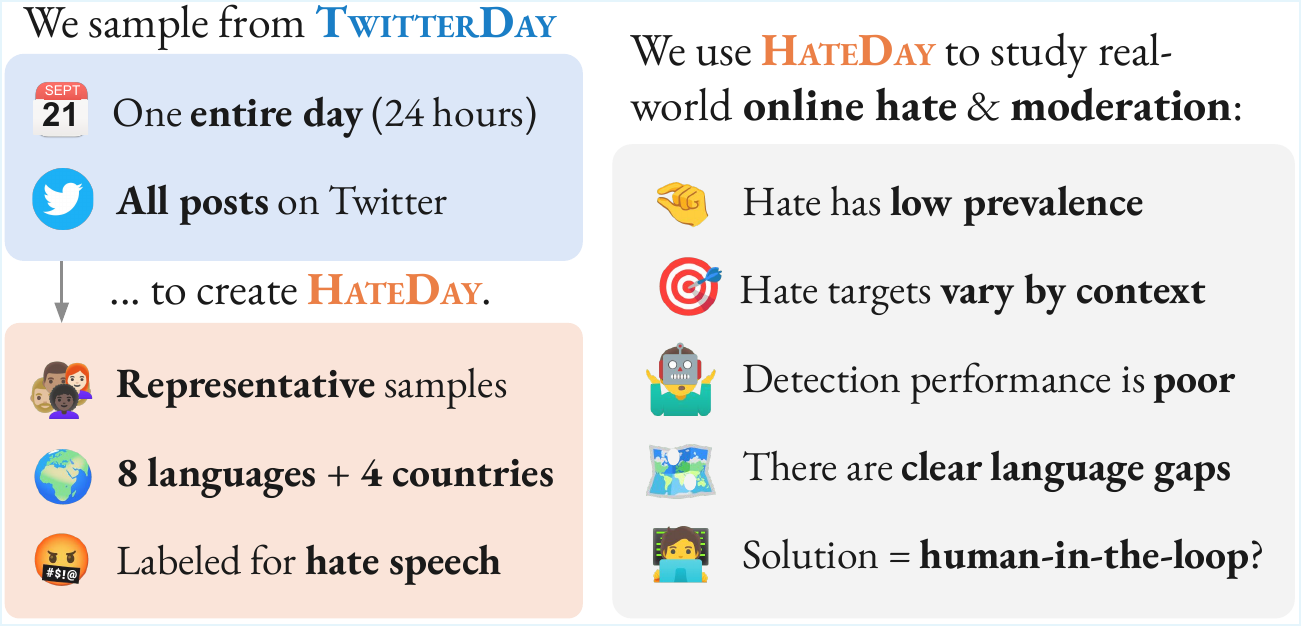} 
    \caption{\textbf{\textsc{HateDay}} consists of twelve annotated representative sets (N=20K each) randomly sampled from all tweets posted on September 21, 2022. The dataset covers eight languages (Arabic, English, French, German, Indonesian, Portuguese, Spanish, and Turkish) and four countries where English is the main language on Twitter (India, Kenya, Nigeria, and the United States). }
    \label{fig:first_page}
\end{figure} 

First, the performance of hate speech detection systems in real-world social media settings remains largely unknown. Industry models are typically not publicly available, and platform transparency reports omit detailed performance metrics or present them in ways that can be misleading or difficult to interpret \cite{giansiracusa2021facebook}. Meanwhile, academic evaluations rely on biased datasets that diverge considerably from real-world distributions, especially regarding class imbalance and topic diversity \cite{nejadgholi-kiritchenko-2020-cross}. This raises concerns that reported performance may be substantially overestimated and not generalize well outside such controlled settings \cite{arango2019hate, wiegand-etal-2019-detection}.
% First, the performance of hate speech detection systems in real-world settings remains largely unknown. Industry models are typically not publicly available, and platform transparency reports often omit performance metrics or present them in misleading ways \cite{giansiracusa2021facebook}. Meanwhile, academic evaluations rely on biased datasets that diverge considerably from real-world distributions in terms of class imbalance and topic distribution \cite{nejadgholi-kiritchenko-2020-cross}, raising concerns that reported performance may be substantially overestimated \cite{arango2019hate, wiegand-etal-2019-detection}. 

The second key limitation of academic research on automatic hate speech detection lies in its narrow language focus. Indeed, most available datasets and detection models have been developed for English \cite{poletto2021resources, tonneau-etal-2024-languages}. While such unequal resource allocation is likely to result in unequal performance across languages, the widespread use of customized datasets for evaluation hinders meaningful cross-lingual comparisons.
% The second key limitation of academic research on automatic hate speech detection lies in its narrow linguistic focus, as the majority of available resources have been developed primarily for English \cite{poletto2021resources, tonneau-etal-2024-languages}. Furthermore, the widespread use of language-specific and highly customized evaluation datasets hinders meaningful cross-lingual comparisons.

Finally, focusing on languages in hate speech detection can hide performance differences between countries sharing a language \cite{ghosh-etal-2021-detecting}. Although recent work develops datasets based on geography rather than language \cite{maronikolakis-etal-2022-listening,castillo-lopez-etal-2023-analyzing}, cross-country performance variations remain unclear. Addressing this limitation requires evaluation datasets providing a comparable evaluation setting across countries with a common language.
% Finally, the focus on languages in the development of hate speech detection systems may obscure performance disparities between countries that share the same language \cite{ghosh-etal-2021-detecting}. Although recent work has begun to develop datasets anchored in geographic context rather than language alone \cite{maronikolakis-etal-2022-listening,castillo-lopez-etal-2023-analyzing}, cross-country performance variations remain unclear. Addressing this gap requires multi-lingual evaluation datasets that are both representative of real-world usage and designed to support meaningful comparisons across languages and countries sharing a common language.

% I'd expect to see a mention of multimodality in a list of problems here. Even if we're not solving / evaluating multimodal approaches, the reality is a perfect model for text would cover only a small potion of the real hate that's online. See https://aclanthology.org/2021.findings-acl.166/

In this article, we address these three limitations by evaluating the performance of state-of-the-art hate speech detection models on real-world social media data across multiple languages and countries. We introduce \textsc{HateDay}, a global Twitter dataset which consists of 240,000 tweets, randomly sampled from all tweets posted on September 21, 2022 and annotated for hate speech (Figure \ref{fig:first_page}, left). Specifically, we sample 20,000 tweets each for eight widely used languages on Twitter: Arabic, English, French, German, Indonesian, Portuguese, Spanish, and Turkish. We also sample 20,000 tweets each for four English-speaking countries with a sizable Twitter presence: India, Kenya, Nigeria, and the United States. 

Using \textsc{HateDay}, we provide insights on real-world online hate (Figure \ref{fig:first_page}, right). We show that the prevalence and composition of hate vary significantly across languages and countries. We then evaluate publicly available hate speech detection models across this global landscape. Our findings reveal that detection performance is substantially overestimated when assessed on standard academic datasets and is strikingly low in real-world social media settings---especially for non-European languages. We identify several factors contributing to this poor performance, including models' limited ability to distinguish hate from offensive speech, and a mismatch between the target categories prevalent in academic datasets and those prevalent in real-world discourse. 

Given these results, we examine the feasibility of using public models for hate speech moderation. We argue that automatic moderation---where posts flagged as hateful are moderated without human oversight---is currently too error-prone to be a viable approach. In contrast, we find that human-in-the-loop systems, in which flagged content is reviewed by moderators, can achieve higher moderation rates across languages and countries, albeit at the cost of substantial human review. 

In sum, our contributions are:
\begin{enumerate}
\itemsep=0em
    \item \textsc{HateDay}, the first global hate speech dataset representative of social media settings, composed of 240,000 annotated tweets covering eight languages and four countries\footnote{Available at \url{https://huggingface.co/datasets/manueltonneau/hateday}}
    %The dataset is not representative of Twitter as a whole since we stratify by language and limit to the top 8
    \item a cross-lingual and cross-national comparison of hate prevalence and composition
    \item an evaluation of real-world detection performance across languages and countries, along with a detailed analysis of the limitations of current detection models
    \item an assessment of the feasibility of hate speech moderation using publicly available models, including a comparison of automatic and human-in-the-loop approaches
\end{enumerate}

\section{The \textsc{HateDay} Dataset}

\begin{figure*}[ht]
    \raggedleft
    \includegraphics[width=1\textwidth]{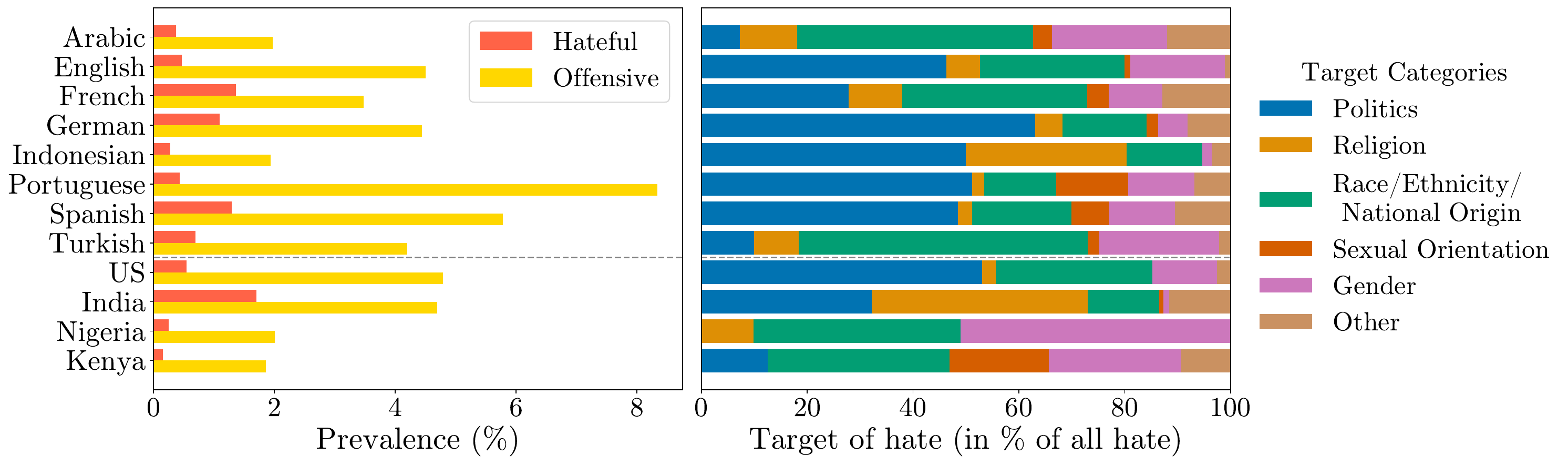} 
    \caption{\textbf{Prevalence of harmful content (left) and targets of hate speech (right) in \textsc{HateDay}}. 
    % Numbers on the left plot bars indicate the number of hateful examples for each language and country. 
    The target category ``Other'' contains rare target labels such as ``Caste'', ``Age'', ``Occupation'', ``Disability'' and ``Social Class''.}
    \label{fig:descriptive_stats}
\end{figure*} 

%%%%%%%%%%%%%%%%%%%%%%%%%%%%%%%%%%%%%%%%%%%%%%%%%%%%%%%%%%%%%%%%%%%%%%%%
\subsection{Data Collection}
As a basis for creating \textsc{HateDay}, we use the \textsc{TwitterDay} dataset \cite{pfeffer2023just}, which contains all Twitter posts within a 24-hour period starting on September 21, 2022.
This amounts to approximately 375 million tweets posted by 40 million users platform-wide.

We filter \textsc{TwitterDay} both at the language and country levels.
At the language level, we focus on the eight most popular languages in the \textsc{TwitterDay} dataset for which there exist hate speech detection resources (details in \S\ref{app:twitterday_lang_distribution}), namely Arabic, English, French, German, Indonesian, Portuguese, Spanish and Turkish.
To assess differences between countries with a common language, we also filter at the country-level, focusing on four countries for which English is the main language on Twitter, namely India, Kenya, Nigeria, and the United States. 
We use the Google Geocoding API to infer user country location based on their self-provided profile location, knowing this has limitations \cite{hecht2011tweets}.
% We provide more details on the language distribution in the dataset and within each country in \S \todo{XXX}.
We drop retweets to focus on original content and randomly sample 20,000 tweets for each of the eight retained languages and each of the four countries, ensuring that each language-specific and country-specific sample is representative of real-world Twitter settings in that particular language or country.
The \textsc{HateDay} dataset corresponds to the combination of all annotated random samples for each language and country, totaling 240,000 annotated tweets. 

% collected 10 minutes after they were posted. Add details about this dataset. We restrict the analysis to six languages and four countries (explain how we get location). Data collected 10 minutes after being posted, so likelihood of moderation is low.

%%%%%%%%%%%%%%%%%%%%%%%%%%%%%%%%%%%%%%%%%%%%%%%%%%%%%%%%%%%%%%%%%%%%%%%%
\subsection{Data Annotation}

We recruit a team of 36 annotators, three per language or country. For languages that are spoken widely around the globe, such as English, Arabic or French, we maximize the diversity of annotator origins as much as possible (details in \S \ref{app: annotator demographics}). 

We follow a \textit{prescriptive} approach to manage annotator subjectivity in our annotation task \cite{rottger-etal-2022-two} by instructing annotators to strictly adhere to extensive annotation guidelines that describe our taxonomy of hate speech (provided in \S \ref{app:annotation_guidelines}). Following prior work \cite{davidsonetal2017, mathew2021hatexplain}, the annotation consists of assigning each tweet to one of three classes:\ (i) \textit{hateful}, if it promotes violence or contains a direct attack, abuse or threat against an individual or a group based on the perceived belonging of a certain characteristic (e.g., gender, race), (ii) \textit{offensive}, if it contains potentially objectionable language that is not hateful, including insults, threats, and posts containing profanity or swear words \cite{zampieri-etal-2019-predicting}, or (iii) \textit{neutral} if it is neither hateful nor offensive. For tweets labeled as hateful, we also ask annotators to specify the groups being targeted as a free-text label. We provide examples of tweets for each class in Table~\ref{tab:hate_examples}. 

\begin{table}[hbt!]
    \centering
    \renewcommand{\arraystretch}{1.2}
    \resizebox{0.48\textwidth}{!}{
    \begin{tabular}{llll}
        \hline
        \textbf{Class} & \textbf{Target Category} & \textbf{Examples} & \textbf{Split} \\
        \hline

        & Politics & \textit{Greens are vermin} & German \\
        \cline{2-4}
        & National Origin &  \textit{Venezuelans are useless} & Spanish\\
        \cline{2-4}
        \textbf{\colorbox{tomato}{Hateful}} & Gender & \textit{I hate women} & Portuguese \\
        \cline{2-4}
        & Religion & Hunt these muslim b*stards & India \\
        \cline{2-4}
         & Sexual Orientation & \textit{F*ck these f*ggots} & French \\
        \hline
        \textbf{\colorbox{gold}{Offensive}} & n/a & Stop spewing rubbish, fool. & Nigeria\\ \hline
        \textbf{\colorbox{darkgrey}{Neutral}} & n/a & \textit{Reunited by the mercy of God} & Turkish \\ \hline 
    \end{tabular}}
    \caption{\textbf{Examples of tweets for each class and main target categories}. Offensive tweets, by definition, have no target. English translations are displayed in italic.}
    \label{tab:hate_examples}

\end{table}

For each language and country, we conduct a pilot annotation phase to train annotators on the task. Specifically, we have each annotator label 100 posts, sampled from a collection of all annotated hate speech datasets \cite{tonneau-etal-2024-languages}, which have been open-sourced on Hugging Face. We repeat this task until the inter-annotator agreement, measured by Krippendorff's $\alpha$, reaches a threshold of 0.7. After completing the pilot, we then task each annotator with labeling the random sample of 20,000 tweets in their respective language or country. Each tweet is labeled by three annotators, and the final label is determined by a simple majority vote.
Across all languages and countries, the three annotators agreed on 93.4\% of all labeled tweets; two out of three agreed in a further 6.4\% of cases, and all three disagreed in only 0.2\% of cases.
% For each language and country, we conduct a pilot annotation to train annotators on the task. Specifically, we have each annotator label 100 posts sampled from a collection of all annotated hate speech datasets \cite{tonneau-etal-2024-languages} open-sourced on Hugging Face. We repeat this task until the inter-annotator agreement, measured by Krippendorff's $\alpha$, reaches 0.7. After the pilot, we task each annotator to label the random sample of 20,000 tweets in their respective language or country. Each tweet is labeled by three annotators, and the final label is determined by majority vote. 
% % To minimize the amount of hate that is wrongly categorized as not hate, all tweets without a majority vote or with at least one hateful label out of three are re-examined by all three annotators and a task expert to derive their final label. 
% Across all languages and countries, the three annotators agreed on 93.4\% of all labeled tweets, two out of three agreed in 6.4\% of cases, and all three disagreed in 0.2\% of cases.
% % (Krippendorff's $\alpha$ = \todo{XX}).
%  % We provide more fine-grained agreement rates in \todo{\S\ref{app:agreement_rates}}. 

%%%%%%%%%%%%%%%%%%%%%%%%%%%%%%%%%%%%%%%%%%%%%%%%%%%%%%%%%%%%%%%%%%%%%%%%
\subsection{Descriptive Statistics}

% We provide descriptive statistics on the prevalence of harmful content and the target composition of hate speech in \textsc{HateDay} in Figure~\ref{fig:descriptive_stats}. 

\paragraph{Prevalence of harmful content} We find that the prevalence of hate speech on Twitter on the day of analysis is very low, representing less than 2\% of all posts across languages and countries, with an average prevalence of 0.7\% (Figure~\ref{fig:descriptive_stats}, left).
In contrast, offensive content is substantially more prevalent than hate speech across all considered languages and countries, ranging from 2.5 times more prevalent for French to 19 times more prevalent for Portuguese. We also find notable differences in the prevalence of hate speech between languages and countries. For instance, 1.0\% of German tweets are hateful whereas only 0.3\% of Indonesian tweets are hateful.  We observe similar gaps at the country level, with the share of hateful posts being much higher in India—1.7\%—compared to 0.2\% in Kenya and 0.3\% in Nigeria.

\paragraph{Main targets of hate speech} We find that the most common targets of hate in \textsc{HateDay} are political as well as racial and gender groups (Figure~\ref{fig:descriptive_stats}, right). We also find notable differences across languages and countries. For instance, political hate speech is prevalent in English, German, Indonesian, Portuguese, and Spanish, as well as in the US national context, representing up to 66\% of all hate in the German context. In contrast, it is less present in Turkish and Nigerian tweets, and not present at all in Arabic tweets. Also, religious hate speech represents 41\% of all hate in India, mostly in the form of Islamophobia, whereas it is less prominent elsewhere and completely absent from the Kenyan sample. Finally, we note that some forms of hate speech are unique to specific contexts, such as casteism in India.

% We observe that \todo{prevalence of hate speech on Twitter on the day of our study is very low, ranging from X to Y\%. large differences between countries in terms of prevalence. In terms of composition, politics and race/ethnicity/national origin dominate but also important differences between languages and countries. Politics dominate in English, in the US, and in German but is less important for French, Turkish or in Nigeria. Religion represents the majority of hate speech in India and a non-trivial share in French and Nigeria, mostly composed of islamophobia in these languages and countries, but less important elsewhere. Also context-specific hate like casteism in India. }

%%%%%%%%%%%%%%%%%%%%%%%%%%%%%%%%%%%%%%%%%%%%%%%%%%%%%%%%%%%%%%%%%%%%%%%%%%%%%%%%%%%%%%
%%%%%%%%%%%%%%%%%%%%%%%%%%%%%%%%%%%%%%%%%%%%%%%%%%%%%%%%%%%%%%%%%%%%%%%%%%%%%%%%%%%%%%

\section{Experimental Setup}

Across our experiments, we leverage the language and country-level representativeness of the \textsc{HateDay} dataset to evaluate public hate speech detection models in real-world social media settings. 
% in real-world social media settings across our languages and countries of interest. 

%%%%%%%%%%%%%%%%%%%%%%%%%%%%%%%%%%%%%%%%%%%%%%%%%%%%%%%%%%%%%%%%%%%%%%%%
\subsection{Models} 

We evaluate models that are either trained for the task of hate speech detection (supervised learning) or prompted for this task (zero-shot learning). 

\paragraph{Supervised learning}
For each language and country, we identify and evaluate all hate speech detection models that are open-sourced on Hugging Face, and trained using supervised learning. In cases where there are more than five models, we limit our analysis to the five most downloaded models on Hugging Face at the time of analysis (August 2024). We provide the full list of open-source models that we evaluated in \S\ref{app:hf-models}. Additionally, we include the Perspective API \cite{lees2022new}, a widely used toxic language detection system also based on supervised learning. Specifically, we use the API’s ``Identity Attack'' attribute, as it most closely aligns with our definition of hate speech.

\paragraph{Zero-shot learning}
We use Aya23 8B \cite{aryabumi2024aya} and Llama3.1 8B \cite{dubey2024llama} for zero-shot learning. We do so because Aya is designed to be multilingual, and Llama3.1 is one of the most capable open-source models available at the time of our analysis. 
% We note that Llama3.1 is instruction tuned for five of the eight languages we investigate. 
We use small model sizes due to compute constraints.
For the zero-shot prompt, see \S \ref{app:zero-shot}.
% cite: https://ai.meta.com/research/publications/the-llama-3-herd-of-models/
%%Llama3.1 is fine-tuned in  German, French, Italian, Portuguese, Hindi, Spanish, and Thai, 
%%Our langs are Arabic, English, French, German, Indonesian, Portuguese, Spanish and Turkish

% \paragraph{In-domain supervised learning} For each language/country, where available, one general-purpose model, one model specifically pretrained on social media data and one multilingual model pretrained on social media data (XLM-T). Training data from \citet{tonneau-etal-2024-languages} for languages; for countries, explain how we build the datasets. Details on train-test split, hyperparameter tuning. 

%%%%%%%%%%%%%%%%%%%%%%%%%%%%%%%%%%%%%%%%%%%%%%%%%%%%%%%%%%%%%%%%%%%%%%%%
\subsection{Evaluation} 

We evaluate models on the \textsc{HateDay} (HD) datasets to estimate real-world detection performance. 
For comparison, we also test models in two traditional evaluation settings, namely on academic hate speech datasets and functional tests.

\paragraph{Academic hate speech datasets (AD)} We measure performance on academic hate speech datasets to understand how results from past work generalize to more realistic settings. We rely on supersets combining all existing hate speech datasets for all eight languages of interest \cite{tonneau-etal-2024-languages}. Given that English hate speech datasets mostly originate from the US \cite{tonneau-etal-2024-languages}, we use English hate speech datasets both for evaluating English- and US-centered models. 
In the absence of supersets for India, Nigeria and Kenya, we survey all existing datasets for each country and combine them to build the supersets (details in \S \ref{app:data_survey}). We restrict the evaluation to a 10\% random sample of all academic  datasets for each language and country to limit inference costs (details in \S \ref{app:language_supersets}).

\paragraph{Functional tests (HC)} We measure performance on functional tests to estimate the ability of models to handle known challenges in hate speech detection. \hspace{-0.8em} We use HateCheck \cite{rottger-etal-2021-hatecheck, rottger-etal-2022-multilingual}, a suite of functional tests for hate speech detection models, covering six of eight languages in \textsc{HateDay}, missing Indonesian and Turkish.

\paragraph{Evaluation metric}
We evaluate model performance using average precision, which corresponds to the area under the precision-recall curve, and is well suited when class imbalance is high.

% \input{tables/table_test}

%%%%%%%%%%%%%%%%%%%%%%%%%%%%%%%%%%%%%%%%%%%%%%%%%%%%%%%%%%%%%%%%%%%%%%%%%%%%%%%%%%%%%%
%%%%%%%%%%%%%%%%%%%%%%%%%%%%%%%%%%%%%%%%%%%%%%%%%%%%%%%%%%%%%%%%%%%%%%%%%%%%%%%%%%%%%%
\section{Results}

%%%%%%%%%%%%%%%%%%%%%%%%%%%%%%%%%%%%%%%%%%%%%%%%%%%%%%%%%%%%%%%%%%%%%%%%
\subsection{Detection Performance}\label{sec_detection_performance}

\begin{table*}[ht]
\centering
\resizebox{\textwidth}{!}{
\begin{tabular}{c|cccccccccccccccccccccccc}
\toprule
\toprule 
 \multirow{3}{*}{\textbf{Model Type}} & \multicolumn{3}{c}{\textbf{Arabic}} & \multicolumn{3}{c}{\textbf{English}} & \multicolumn{3}{c}{\textbf{French}} & \multicolumn{3}{c}{\textbf{German}} & \multicolumn{3}{c}{\textbf{Indonesian}} & \multicolumn{3}{c}{\textbf{Portuguese}} & \multicolumn{3}{c}{\textbf{Spanish}} & \multicolumn{3}{c}{\textbf{Turkish}} \\
 & \holdout AD & \hatecheck HC & \hateday HD & \holdout AD & \hatecheck HC & \hateday HD & \holdout AD & \hatecheck HC & \hateday HD & \holdout AD & \hatecheck HC & \hateday HD & \holdout AD & \hatecheck HC & \hateday HD & \holdout AD & \hatecheck HC & \hateday HD & \holdout AD & \hatecheck HC & \hateday HD & \holdout AD & \hatecheck HC & \hateday HD \\
\midrule
 Llama 3.1 & \holdout 10 & \hatecheck 76.4 &  \hateday 1.2 & \holdout 38.8 & \hatecheck 88.5 &  \hateday 4.6 & \holdout 33.5 &  \hatecheck 84.9 &  \hateday 7.3  &  \holdout 18.9 & \hatecheck 84.2 &  \hateday 5.8 &  \holdout 62.8 & \hatecheck - &  \hateday 2.5 & \holdout 15.5  & \hatecheck 84.8 &  \hateday 2.2 &  \holdout 39.4 &  \hatecheck 84.9 &  \hateday 7.2 &  \holdout 39.3 & \hatecheck - &  \hateday 2.9 \\
  Aya 23 &  \holdout 11.5 & \hatecheck 77.6 &  \hateday 2 & \holdout 36.1 & \hatecheck 87.4 & \hateday 3.5 & \holdout 37.1 & \hatecheck 84 & \hateday 5.3 & \holdout 21.6 & \hatecheck 86 & \hateday 5.9 & \holdout 61.6 & \hatecheck - & \hateday 2.6 & \holdout 15.9 & \hatecheck 86.6 & \hateday 2.2 & \holdout 40.7 & \hatecheck 85.3 & \hateday 7.6 & \holdout \textbf{44.6} & \hatecheck - &  \hateday 4.5 \\
\midrule
 Best OS & \holdout \textbf{25.8} & \hatecheck 76.9 & \hateday  5.8 & \holdout 38.2 & \hatecheck 82.2 & \hateday 9 & \holdout 35.7 & \hatecheck 89.9 & \hateday 16 &  \holdout \textbf{64.3} & \hatecheck \textbf{97.2} &  \hateday \textbf{19.3} & \holdout \textbf{90.2} & \hatecheck - & \hateday 3.5 & \holdout 18.7 & \hatecheck 85.8 & \hateday 3.1 &  \holdout \textbf{54.9} & \hatecheck 82.3 & \hateday 11.6 &  \holdout 32.1 & \hatecheck - & \hateday 1.9\\
  Perspective & \holdout 18.7 &  \hatecheck \textbf{89.6} &  \hateday \textbf{10.2} & \holdout \textbf{58.9} &  \hatecheck \textbf{95.1} & \hateday \textbf{10.1} & \holdout \textbf{40.1} & \hatecheck \textbf{96.9} & \hateday \textbf{37.7} &  \holdout 53.6 &  \hatecheck 96 & \hateday 18.8 &  \holdout 65.5 & \hatecheck- &  \hateday \textbf{11.1}  & \holdout \textbf{30.9} &  \hatecheck \textbf{94.6} & \hateday \textbf{14.5} &  \holdout 50.9 &  \hatecheck \textbf{96} & \hateday \textbf{34.1} & \holdout 31.4\textsuperscript{*} & \hatecheck - &   \hateday \textbf{6.5}\textsuperscript{*}\\
  % $1.7^{*}$
% \midrule
%   Monolingual, GP & & & & & & & & &  & XX.X & XX.X & 72.0 & 7.9 & 83.8 & 14.9 &  &  & 57.6 & 5.9 & 64.5 & 7.6 & - & XX.X \\
%  Monolingual, SM & & & & & & & & &  & XX.X & XX.X & 78.2 & 7.6 & - & - &  &  & 52.0 & 5.3 & 66.0 & 8.6 & - & XX.X \\
%   Multilingual, SM & & & & & & & & &  & XX.X & XX.X & 74.0 & 5.7 & 81.8 & 16.5 &  &  & 59.2 & 5.3 & 70.9 & 6.5 & - & XX.X \\    
\bottomrule
\bottomrule
\end{tabular}
}
    \caption{\textbf{Model performance across languages and evaluation sets}, as measured by average precision (\%). We report performance on three evaluation sets: \colorbox{holdout}{academic hate speech datasets (AD)} combined for a given language, \colorbox{hatecheck}{HateCheck functional tests (HC)} and \colorbox{hateday}{\textsc{HateDay}} \colorbox{hateday}{(HD)}. HC does not cover Indonesian and Turkish. Asterisks indicate that the language of interest is not supported by the Perspective API.}
    \label{tab:main_results_language}
\end{table*}

We report the average precision of each model across the three aforementioned datasets, across languages (Table~\ref{tab:main_results_language}) and across countries (Table~\ref{tab:main_results_country}). We also provide results on individual open-source model performance in \S\ref{app:indiv_model_perf}.

\paragraph{Performance across datasets}  Our most striking finding is that across the wide range of models, languages, and countries considered, detection performance is much lower on our representative \textsc{HateDay} dataset (HD) than on academic hate speech datasets (AD) or functional tests (HC). Indeed, average precision is just 9.4\% on HD, compared to 40\% on AD and 87.2\% on HC.

% Our most striking finding is that across the wide range of models, languages and countries considered in this study, detection performance is very low on our representative \textsc{HateDay} dataset (HD) compared to performance on academic hate speech datasets (AD) and on the functional tests (HC). Indeed, average precision equals just 9.4\% on average on HD, compared to 40\% on AD and 87.2\% on HC. 
% At the language level, this performance gap is particularly large between HateCheck, the suite of functional tests on which most models perform very well, and \textsc{HateDay}.

% This emphasizes our finding that models that are considered high-performing using traditional evaluation standards still tend to perform poorly in real-world settings. 

% This finding highlights the risk of considerably overestimating classification performance when evaluating hate speech detection models on a dataset whose
% characteristics greatly differ from real-world conditions. 

\paragraph{Performance across model types} We find that supervised learning consistently outperforms zero-shot learning across almost all combinations of language, country, and evaluation sets. Indeed, the best open-source model on \textsc{HateDay} has an average precision of 41.1\% on average across all combinations of language or country and dataset, whereas Aya23 8B, which performs on par with Llama3.1 8B, scores 32.4\% on average. Additionally, despite being originally developed to detect toxic rather than hateful content, we observe that the Perspective API (44.3\% average precision on average) outperforms open-source hate speech detection models in all languages except German. 
At the country-level, open-source models perform best in Nigeria, but Perspective API has higher performance in the US and India. We did not find open-source supervised models to compare Perpective API to in the Kenyan context.
%Still, the dominance of Perspective is more limited at the country-level where it performs best in the US and Indian contexts but not in the Nigerian context, and only in the Kenyan context because there are no other supervised models to compare to. 

\begin{table}[h]
\centering
\resizebox{0.48\textwidth}{!}{
\begin{tabular}{c|cccccccc}
\toprule
\toprule 
 \multirow{3}{*}{\textbf{Model Type}} & \multicolumn{2}{c}{\textbf{United States}} & \multicolumn{2}{c}{\textbf{India}} & \multicolumn{2}{c}{\textbf{Nigeria}} & \multicolumn{2}{c}{\textbf{Kenya}}  \\
 & \holdout AD & \hateday HD & \holdout AD & \hateday HD & \holdout AD & \hateday HD & \holdout AD & \hateday HD  \\
\midrule
 Llama 3.1 & \holdout 38.8 & \hateday 4.9 & \holdout 50.9 & \hateday 13.4 & \holdout 32 & \hateday 2.6 & \holdout 24.3 & \hateday 1.5 \\
  Aya 23 & \holdout 36.1 & \hateday 4.7 & \holdout 52.7 & \hateday 10.4 & \holdout 30.7 & \hateday 1.6 & \holdout 23.9 &  \hateday 0.8 \\
\midrule
 Best OS & \holdout 38.2 & \hateday 9.8 & \holdout 54.3 & \hateday 7.8 & \holdout  \textbf{65.7} & \hateday \textbf{30.9} & \holdout - & \hateday - \\
  Perspective & \holdout \textbf{58.9} & \hateday \textbf{12.3} & \holdout \textbf{61.9} & \hateday \textbf{42.9} & \holdout 44.1 & \hateday 8.6 & \holdout \textbf{31.6} &  \hateday \textbf{9.1} \\
  % $1.7^{*}$
 
\bottomrule
\bottomrule
\end{tabular}
}
    \caption{\textbf{Model performance across countries and evaluation sets}, as measured by average precision (\%). There are no open-source models specifically for Kenya.}
    \label{tab:main_results_country}
\end{table}

\paragraph{Cross-lingual and geographic gaps} We observe substantial performance differences in the real-world setting of \textsc{HateDay} across languages and countries. At the language level, average precision is higher for European languages---English, French, Spanish, Portuguese, and German---averaging 23.1\%---than for non-European languages---Arabic, Indonesian, and Turkish---at 9.3\% on average. At the country level, performance is highest on Indian tweets (42.9\%), followed by Nigeria (30.9\%), and lower in both the US (12.3\%) and Kenya (9.1\%).

% We observe substantial performance differences in the real-world setting of \textsc{HateDay} across languages and countries. At the language-level, we find that the best average precision for European languages (23.1\% on average), namely English, French, Spanish, Portuguese and German, is substantially higher than for non-European languages (9.3\%), namely Arabic, Indonesian and Turkish. At the country-level, the average precision is by far highest on Indian tweets (42.9\%) followed by Nigeria (30.9\%), whereas the performance is rather low in both the US\ (12.3\%). and Kenyan (9.1\%) contexts. 

%%%%%%%%%%%%%%%%%%%%%%%%%%%%%%%%%%%%%%%%%%%%%%%%%%%%%%%%%%%%%%%%%%%%%%%%
\subsection{Reasons for Low Performance}

In the following, we conduct a more in-depth analysis into potential explanations for the poor hate speech detection performance as well as the cross-geographic performance gaps we observed in real-world settings in \S\ref{sec_detection_performance}.

\paragraph{Low precision and recall} As the average precision corresponds to the area under the precision-recall curve, the same average precision may correspond to very different patterns in precision and recall. We therefore inspect the precision-recall curves of the best-performing models on \textsc{HateDay} for each language and country (Figure \ref{fig:precision_recall} in the Appendix). We find that, while precision or recall may be higher depending on the context, both values remain very low and there are no situations where both values are above 50\%, with the highest F1-score at 47\% for India. This highlights the role of both false positives and negatives in the observed low performance. Next, we study the composition of these two types of errors.

\begin{figure}[ht]
    \raggedleft
    \includegraphics[width=0.5\textwidth]{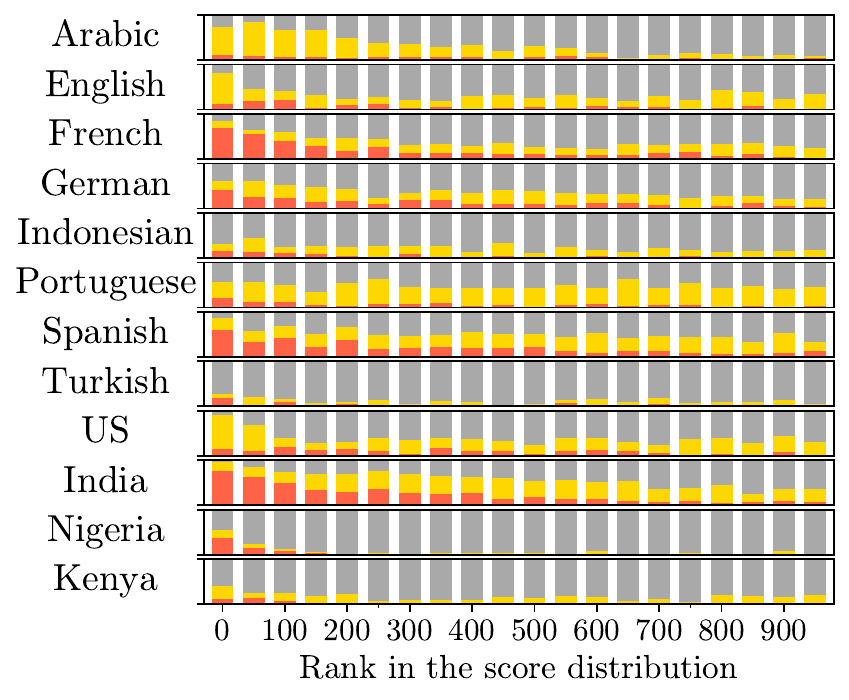} 
    % \caption{Share of hateful, offensive and neutral in the top 5\% (N=1,000) of the score distribution of the best model on \textsc{HateDay} for each language and country. }
    \caption{\textbf{Share of \colorbox{tomato}{hateful}, \colorbox{gold}{offensive} and \colorbox{darkgrey}{neutral} content} in the top 5\% scored tweets (N=1,000) in \textsc{HateDay} for each language and country. We use the hatefulness score distribution of the best performing model on \textsc{HateDay} (Tables~\ref{tab:main_results_language} and \ref{tab:main_results_country}). The x-axis corresponds to the descending rank and each bar shows the distribution of content in a window of 50 tweets.}
    \label{fig:share_content_across_ranks}
\end{figure}

\paragraph{Offensive false positives}  
We find that offensive content constitutes a substantial share of the top of the hatefulness score distribution---32.3\% of the top 50 (0.25\%) scored tweets on average---for each language and country (Figure \ref{fig:share_content_across_ranks}), thereby significantly hurting precision. For Arabic, English, Portuguese, and the US and Kenyan national contexts, offensive content even appears more frequently than hate speech at the very top of the score distribution. This problem is further aggravated by the fact that offensive content is more prevalent than hate speech (Figure \ref{fig:descriptive_stats}). Indeed, we observe that as the prevalence ratio of offensive content to hate speech steadily increases, the share of offensive content at the top of the score distribution also gradually increases (Figure \ref{fig:increasing_offensive_ratio} in the Appendix), exacerbating the negative impact on precision.

% \paragraph{Offensive false positives}  We find that offensive content constitutes a substantial share of the top of the hatefulness score distribution --- 32.3\% in the top 50 (0.25\%) scored tweets on average --- for each language and country (Figure \ref{fig:share_content_across_ranks}), thereby hurting precision. For Arabic, English, Portuguese and the US\ and Kenyan national contexts, offensive content even appears more frequently than hate speech at the top of the score distribution. This problem is further aggravated by the fact that offensive content is more prevalent than hate speech on the day of the analysis (Figure \ref{fig:descriptive_stats}). Indeed, we observe that as the prevalence ratio of offensive content to hate speech increases, the share of offensive content also gradually increases (Figure \ref{fig:increasing_offensive_ratio} in the Appendix).  
\vspace{-0.5mm}
\paragraph{Qualitative analysis of false positives} We further examine top-scored neutral tweets to identify other prominent features of false positives beyond offensiveness. We find that many such tweets contain mentions of hate speech (e.g., ``was it the political scientist saying that Ukrainians are `subhumans'?''). We also find several cases of statements that are not hateful but would be with a few changes of letters (e.g., ``emigration is a catastrophy''). Finally, we find that a substantial share of such tweets, usually replies, may be hateful but lack the context to conclude (e.g., ``@USER illegal migrants, they say it themselves'').

\paragraph{Analysis of false negatives}
Next, we analyze false negatives, defined as hateful examples missing from the top 1\% of the best model's score distribution for each language and country. We first examine the targets of these tweets to compare their representation in false negatives to their overall prevalence in hate speech for each language or country. We find that politics and gender are overrepresented in false negatives, respectively in 10/12 and 7/12 of cases. Conversely, religion and race are underrepresented in 10/12 and 8/12 of cases. We also consider ambiguity, defining a hateful example as ambiguous if not all three annotators labeled it as hateful. We find that the share of ambiguous content is significantly higher in false negatives than in all hateful examples (t(11) = 3.76; $p< 0.05$).

% Next, we analyze false negatives, defined as hateful examples that do not appear in the top 1\% of the best model's score distribution for each language and country. We first examine the targets of these tweets to determine how their representation in false negatives compares to their overall prevalence in hate speech for each specific language or country. We find that politics and gender are overrepresented in false negatives, respectively in 10/12 and 7/12 of the studied language and country-level cases. In parallel, religion and race are under-represented, respectively in 10/12 and 8/12 of cases. We also look at ambiguity, considering that a hateful example is ambiguous if not all three annotators labeled it as hateful. The share of ambiguous content is significantly higher in false negatives compared to the whole set of hateful examples across languages and countries (t(11) = 3.76; $p< 0.05$). 

\paragraph{Target academic focus and prevalence} Motivated by the overrepresentation of certain targets in false negatives, we inspect the role of target-level academic focus, as a proxy for the distribution of hate targets seen by the model during training, on performance. We first examine differences at the target level between the prevalence of each target in academic datasets and its prevalence in \textsc{HateDay} across languages and countries. We use data from a recent survey which documents the share of all hate speech datasets by target category and language \cite{yu2024unseen}. In the absence of data at the country level, we categorize all surveyed datasets in terms of the target focus for each country of interest (details in \S\ref{app:target_categorization}). For each language and country, we provide a comparison between the target-level shares of (i) all hate speech datasets and of (ii) all hate speech in \textsc{HateDay} in Figure \ref{fig:comparison_prevalence_academic} in the Appendix. Our most notable finding is that political hate speech is much more prevalent in \textsc{HateDay} for two-thirds of the 12 languages or countries than in academic datasets. To a lesser extent, we also find that gender-based hate is often more prevalent in \textsc{HateDay} than is studied while religion-based hate is less prevalent than is studied in existing hate speech datasets. 

\paragraph{Target alignment and performance} Finally, we explore the impact of the alignment between the target focus of academic work and the target prevalence in real-world social media data on performance. We estimate such alignment for each language and country by computing the cosine similarity between the vectors containing the share of all hate speech datasets by target focus and the share of all hate speech in \textsc{HateDay} by target. We find a strong and significant positive correlation between this alignment and detection performance (Pearson's \textit{r}:\ 0.76, $p=0.003$). This indicates that the better hate speech detection resources reflect hate target coverage in real-world social media data for a given language or country, the higher the detection performance. In contrast, we do not find a significant correlation between the amount of annotated datapoints in the aforementioned supersets and performance on \textsc{HateDay} (Pearson's \textit{r}:\ -0.47, $p=0.127$). This indicates that the amount of detection resources may play a smaller role in cross-lingual and geographic performance differences compared to the alignment between academic target focus and real-world target prevalence.

% \paragraph{Cross-geographic performance gaps} Finally, we study two potential factors explaining differences in model performance on \textsc{HateDay} across languages and countries: (i) the amount of resources publicly available to train detection models, using the number of annotated datapoints for hate speech detection as a proxy, and (ii) the alignment between the target focus of academic work and the target prevalence in real-world settings. We estimate the latter for each language and country by computing the cosine similarity between the vectors containing the share of all hate speech datasets by target focus and the share of all hate speech in \textsc{HateDay} by target. We find a strong and significant positive correlation between the alignment of target-level academic focus and real-world prevalence on the one hand and detection performance on the other hand (Pearson correlation coefficient: 0.76, p-value: 0.003). This indicates that the better hate speech detection resources reflect hate target coverage in real-world settings for a given language or country, the higher the detection performance. In contrast, we do not find a significant correlation between the amount of annotated datapoints to train models and performance on \textsc{HateDay} (Pearson correlation coefficient: -0.47, p-value: 0.127), indicating that data availability alone cannot explain cross-lingual and geographic performance differences. 

% Role of ambiguity.

%%%%%%%%%%%%%%%%%%%%%%%%%%%%%%%%%%%%%%%%%%%%%%%%%%%%%%%%%%%%%%%%%%%%%%%%
\subsection{Feasibility of Hate Speech Moderation}

\begin{figure}[ht]
    \raggedleft
    \includegraphics[width=0.48\textwidth]{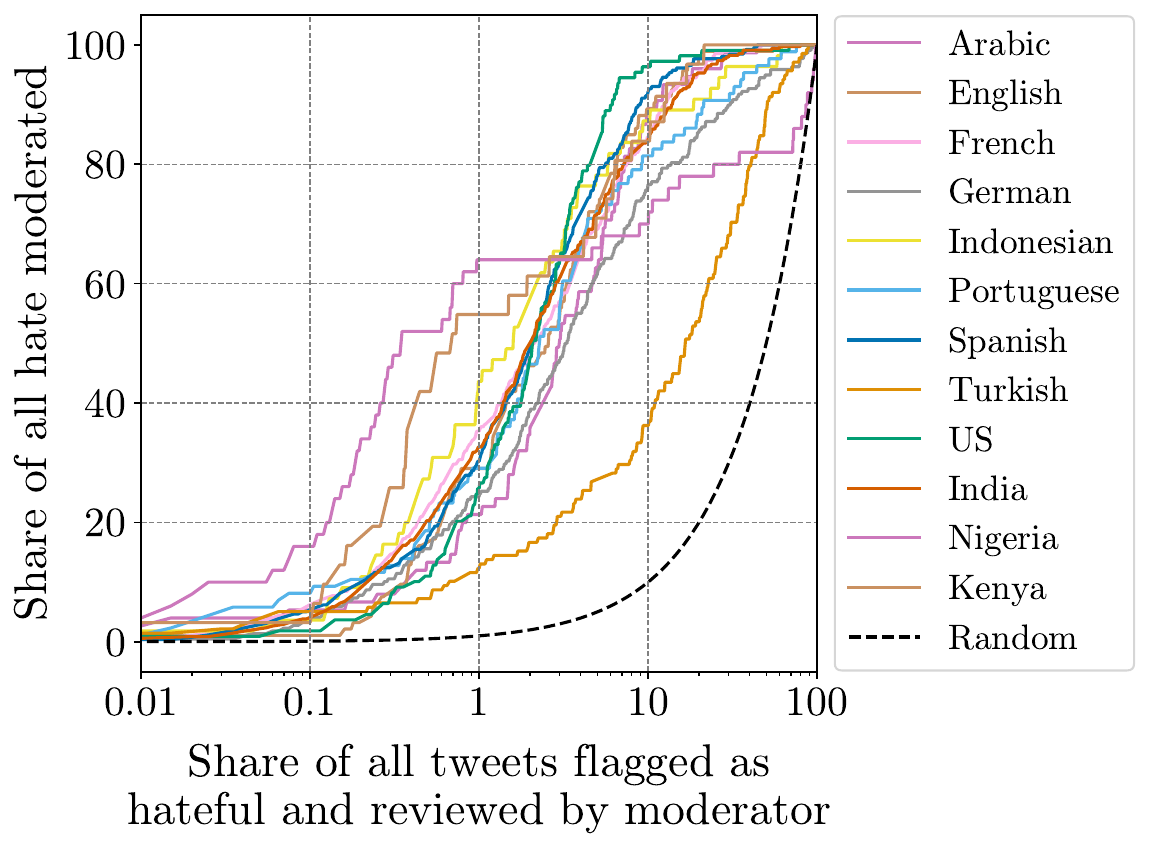} 
    % \caption{Share of hateful, offensive and neutral in the top 5\% (N=1,000) of the score distribution of the best model on \textsc{HateDay} for each language and country. }
    \caption{\textbf{Cost-recall tradeoff in human-in-the-loop moderation}. Share of all \textsc{HateDay} tweets flagged as hateful and reviewed by moderators (\%) versus share of all moderated hate in \textsc{HateDay} (\%). We use the best model on \textsc{HateDay} for each language and country (Tables~\ref{tab:main_results_language} and \ref{tab:main_results_country}). The dashed line indicates performance of a model that flags tweets as hateful at random.}
    \label{fig:human_in_loop_moderation}
\end{figure} 

Most social media platforms prohibit hate speech \cite{singhal2023sok} and have done so since their inception \cite{gillespie2018custodians}. In light of our poor performance results (\S\ref{sec_detection_performance}), we investigate the feasibility of hate speech moderation using publicly available hate speech detection models.
% However, with billions of daily posts worldwide, ensuring consistent enforcement is a formidable challenge. This vast scale has driven the adoption of algorithmic detection methods \cite{gillespie2020content}, 
Given the demonstrated low performance of hate speech detection in real-world social media settings, automatic moderation, whereby content flagged as hateful by a detection model is directly moderated, may be too error-prone. 
We therefore consider human-in-the-loop moderation, which is a more realistic approach used by large social media platforms \cite{avadhanula2022bandits} whereby content likely to be hateful is flagged by detection models for review by human moderators. We study the trade-off between the amount of human reviewing required and the corresponding share of all hate successfully moderated (Figure \ref{fig:human_in_loop_moderation}, details in \S\ref{app:human-in-the-loop}).

We find that successfully moderating a large share of all hateful content (>80\%) using the best publicly available detection models would require human review of at least 10\% of all daily tweets. For such a review workload, the total share of hate successfully moderated varies between 70\% and 90\% for most languages or countries, apart from Turkish where this share is only 40\%. 10\% of all daily tweets may represent a massive amount of posts depending on the context and ranges from 19,300 tweets for Kenya to 5.1 million for English. 

% \todo{[make distinction clearer between feasibility for small communities and prohibitive cost for larger communities]} 
A smaller share of human review would lead to most hate being left unmoderated:\ when flagging and reviewing 1\% of all tweets posted in a given day, only 20-40\% of all hate is moderated for most languages and countries. 
% These results point to the limits of hate speech moderation in its current form.

%%%%%%%%%%%%%%%%%%%%%%%%%%%%%%%%%%%%%%%%%%%%%%%%%%%%%%%%%%%%%%%%%%%%%%%%%%%%%%%%%%%%%%
%%%%%%%%%%%%%%%%%%%%%%%%%%%%%%%%%%%%%%%%%%%%%%%%%%%%%%%%%%%%%%%%%%%%%%%%%%%%%%%%%%%%%%
\section{Discussion}

\paragraph{Hate speech prevalence} We find that the proportion of hate speech relative to all social media content is very low, which is in line with past work \cite{gagliardone2016mechachal,mondal2017measurement, park2022measuring}. This contrasts with recent survey results where two thirds of social media users report that they often encounter hate speech in their feeds \cite{unesco2023survey} and points to the potential role of recommendation algorithms in amplifying harmful content \cite{milli2025engagement}. 
% something about how hate speech/target prevalence varies across countries

\paragraph{Performance overestimation} We find that traditional evaluation of detection models, on academic datasets and functional tests, largely overestimates performance in randomly drawn samples of real-world social media data. Performance on real-world data is very low, especially for non-European languages. 
% We attribute this result to the higher diversity of both positives (e.g., different targets) and negatives (e.g., offensive, ambiguous content) in real-world settings compared to hate speech datasets. \todo{[functional tests weight all tests equally but prevalence of situations differ: marking offensive content as non-hateful should have a high weight]} 
While past work has discussed the risk of such overestimation \cite{arango2019hate}, due to the biased nature of hate speech datasets \cite{wiegand-etal-2019-detection, nejadgholi-kiritchenko-2020-cross}, and quantified such risk in the Nigerian Twitter context \cite{tonneau-etal-2024-naijahate}, we provide evidence of low and overestimated performance across many more linguistic and national contexts. 
% \todo{[shows that hate speech detection problem far from being solved, governments and non-profits unaware \cite{parker2023hate}]}
Moreover, our results highlight the need to evaluate NLP tools in realistic settings, especially where human harm might arise. This applies more broadly to the detection of harmful content on social media, such as misinformation \cite{thorne-etal-2018-fever,magomere2025claims}, as well as the evaluation of the bias and safety of generative large language models \cite{ibrahim2024beyond,lum2024bias, rottger-etal-2024-political,rottger2025issuebench}, both of which have relied so far on non-representative benchmarks for evaluation.

\paragraph{Supervised beats zero-shot} Despite growing enthusiasm for decoder-based large language models (LLMs), we find that they underperform supervised learning for hate speech detection—echoing findings from prior work \cite{nozza-2021-exposing}. This also reinforces the conclusions of a recent survey of NLP practitioners, which highlights the continued importance of annotated data for maximizing model performance \cite{romberg2025have}.

\paragraph{Dominance of Perspective API} We find that the Perspective API often outperforms other models in hate speech detection across languages, contradicting existing evaluations on traditional hate speech datasets, which showed that Perspective is outperformed by academic hate speech classifiers \cite{wich2022introducing,ofcom2024}. We attribute such dominance to the fact that Perspective is optimized for real-world performance and generalizability to unseen data \cite{lees2022new}, contrary to traditional hate speech detection models developed in academia. We also note that the Perspective API is periodically updated and that performance may have improved since the evaluations conducted in past work \cite{pozzobon-etal-2023-challenges}. Still, we find that overall performance remains low, which echoes past evaluations of Perspective's performance and biases \cite{nogara2023toxic,hartmann2025lost}. Also, we observe that Perspective's dominance does not always hold at the country level, with the open-source hate speech classifier outperforming Perspective in the case of Nigeria. This may be due to conceptual alignment: the \textsc{HateDay} annotation guidelines were partly based on the same definition and instructions used to train the Nigerian classifier \cite{tonneau-etal-2024-naijahate}. This also highlights the limits of the language-level approach in developing Perspective and more generally hate speech classifiers \cite{tonneau-etal-2024-languages}, which may not be tailored to national contexts where non-US English and code-mixing is prevalent, as already demonstrated in past work \cite{ghosh-etal-2021-detecting,haber-etal-2023-improving}. 
% \todo{[something on the fact that despite the enthusiasm around decoder-based LLMs for many detection tasks, it doesn not work well for hate speech detection, in line with past work]}

\paragraph{The problem of offensive language} We find that a major reason for the poor performance of hate speech classifiers on \textsc{HateDay} is the high prevalence of offensive content at the top of the hate score distribution, which reduces precision. While offensive content differs from hate speech because it does not target an individual or a group based on the perceived belonging of a particular characteristic (e.g., gender, race), it shares lexical features such as the use of swear words and profanities. These similarities likely explain the high hate scores given by classifiers to offensive content. While prior work initially documented this issue \cite{davidsonetal2017}, our findings demonstrate its heightened relevance in real-world social media settings, as it is aggravated by the substantially higher prevalence of offensive content compared to hate speech.

% \todo{\paragraph{Cross-lingual/country differences} say something about this?}

\paragraph{Unseen targets of hate} Our findings reveal a misalignment between the targets of hate speech studied in academic work and their real-world prevalence in \textsc{HateDay}. Notably, political-related hate is understudied in academic work compared to its prevalence in \textsc{HateDay}, expanding past work that points to a low representation of certain targets in hate speech detection resources \cite{yu2024unseen}. We also find that the alignment between academic target coverage and real-world target prevalence has a significant positive correlation with detection performance in \textsc{HateDay}, contrary to the raw amount of detection resources. Improving such alignment, particularly for underrepresented hate types such as political-based hate, is therefore crucial for enhancing real-world detection performance. In the process, political criticism and hate speech must be clearly differentiated to avoid accusations of politically biased moderation \cite{pew2020politbias} and hate speech laws being misused for political censorship \cite{strossen2018hate}.

\paragraph{Cross-geographic disparities}

Consistent with prior research \cite{rottger-etal-2022-multilingual}, we find that detection performance is generally lower for non-European languages, such as Arabic, Turkish and Indonesian. Our analysis extends this finding by offering one explanation mentioned in the past paragraph: performance disparities may stem less from varying amounts of annotated training data and more from a misalignment between the target categories prevalent in academic datasets and those that are prevalent in real-world content. This insight may also explain counterintuitive results such as the surprisingly low real-world performance for the English and US contexts and higher performance for India, despite English and the US having the most annotated resources \cite{tonneau-etal-2024-languages}. In real-world English-speaking and US contexts, politically motivated hate is among the most prevalent forms of hate speech; yet, it is underrepresented in US-dominated English academic datasets. By contrast, in the Indian context, there is greater alignment between the types of hate emphasized in academic datasets and those commonly encountered in practice, which may help explain the observed performance differences. Furthermore, the overall prevalence of hate speech is significantly lower in English and US data (0.4\% and 0.5\% of posts, respectively) compared to Indian data (1.7\%), suggesting that our estimate of model performance in English and US settings may be less robust.
% Generally, for non-English Global South languages, performance is lower, in line with past work \cite{rottger-etal-2022-multilingual}. Our analysis goes further and provides explanations: performance differences more because of misalignment between academic target focus and real-world target prevalence rather than quantity of annotated resources. complements past work with similar findings with more detailed explanation. This result Might also explain counterintuitive results with low performance for English and a higher performance for India than for US, even though this language/country has the most resources \cite{tonneau-etal-2024-languages}; political-related hate most prevalent in the wild for english/US whereas it's only poorly represented in US-dominated English academic datasets. In contrast, alignment between academic target focus and target prevalence in the wild for India is higher, potentially explaining performance differences. Prevalence of hate for English and US contexts (X and Y\% hateful posts respectively) much lower than for India (Z\%) so English/US results might also be less robust. 

\paragraph{Other challenges for hate speech detection}
We present further challenges for hate speech detection that have partly been identified in past work:\ First, the difficulty of distinguishing between the use and the mention of hate speech, for instance when slurs or hateful terms are quoted as part of counter speech or educational content \cite{rottger-etal-2021-hatecheck, gligoric-etal-2024-nlp, jin2024disentangling}. Second, the inability of classifiers to capture subtle differences in wording and phrasing that can make a non-hateful statement become hateful \cite{sap-etal-2019-risk}. Third, the lack of context in many posts---especially replies---poses challenges to classification and calls for more contextualized hate speech detection \cite{perez2023assessing}. Our analysis underlines the relevance of these challenges, given their prevalence in real-world settings, and highlights the need to address them as a priority.
% These findings are in line with past work and need to be addressed to improve real-world detection performance.

\paragraph{Implications for hate speech moderation} Our results indicate that fully automated hate speech moderation using public hate speech detection models is undesirable. Indeed, low real-world detection performance renders fully automated moderation ineffective and potentially harmful, as it fails to protect users from hate speech while likely removing non-hateful content, such as counter-speech. This reinforces concerns raised in earlier theoretical work \cite{gorwa2020algorithmic} and complements past empirical findings highlighting the limitations of hate speech detection research for content moderation \cite{ye-etal-2023-multilingual,zheng-etal-2024-hatemoderate}.
% \todo{\cite{gomez2024algorithmic}}
We also show that human-in-the-loop moderation based on public detection models can effectively moderate a large share of hate speech across our languages and countries of interest but that this implies having humans review a non-trivial share of all daily content, extending past results for the Nigerian Twitter context \cite{tonneau-etal-2024-naijahate} to more languages and countries. Beyond financial costs, the required scale of human review also raises concerns about potential reviewer harm from repeated exposure to harmful content \cite{roberts2016commercial,kirk-etal-2022-handling}.  
Our results support the claim that detection alone will not solve the hate speech problem \cite{parker2023hate} and call for complementary solutions. These include preventive approaches that aim to reshape speech norms and have proven effective in curbing hate speech, for instance by prompting users to reconsider harmful posts \cite{katsaros2022reconsidering} or by confronting hate speech with counter speech \cite{munger2017tweetment,hangartner2021empathy}.
% Finally, we note that our results do not imply that hate speech moderation itself is ineffective, because social media platforms use their own private models for moderation and have access to more data which may improve performance (see Limitations). Yet, without access to real-world performance metrics for platform models, the effectiveness of hate speech moderation on social media remains uncertain.
% \todo{[adapt to the real world: related to the necessity for classifiers to be in line with moderation rules \cite{zheng-etal-2024-hatemoderate}]}
% Taken together, these findings suggest that current approaches are insufficient for significantly limiting the spread of online hate speech. \todo{necessity to evaluate platform models}
% \todo{would be important to test platform models and see whether they fare better} 

%%%%%%%%%%%%%%%%%%%%%%%%%%%%%%%%%%%%%%%%%%%%%%%%%%%%%%%%%%%%%%%%%%%%%%%%%%%%%%%%%%%%%%
%%%%%%%%%%%%%%%%%%%%%%%%%%%%%%%%%%%%%%%%%%%%%%%%%%%%%%%%%%%%%%%%%%%%%%%%%%%%%%%%%%%%%%

\section{Conclusion}

In this article, we introduced \textsc{HateDay}, the first global hate speech dataset representative of real-world social media settings.  Using \textsc{HateDay}, we showed that evaluating hate speech detection models on standard academic datasets substantially overestimates their real-world performance, which is very low. We explored the implications of this finding for content moderation, concluding that relying on public detection models for automated moderation is currently ill-advised due to their high error rate. Accordingly, improving real-world model performance should be a key focus of future research. We also found that human-in-the-loop moderation can be more accurate, but only if a substantial portion of daily content is manually reviewed---raising important questions about the feasibility and desirability of such an approach at scale, which merits further investigation.
% [remaining question: is hate speech moderation feasible? model gap: encourage platform to be more transparent on detection performance of their content moderation systems]
% In light of our results, future work is needed to improve hate speech detection performance + as whether the required scale of such human review is feasible and desirable remains an open debate and warrants further investigation..
% , for instance by focusing on under-studied types of hate such as political hate. 
Ultimately, we urge researchers to evaluate future detection models within the real-world contexts where they are likely to be deployed. We also call on platforms to provide greater transparency about how their detection systems perform in real social media environments, to better assess moderation effectiveness. We hope that our dataset and findings will drive progress in both areas and contribute to addressing this pressing challenge.
% Overall, we encourage researchers to ground their evaluation in the real-world settings in which their models may eventually be used. We also encourage platforms to be more transparent on how their detection models perform in social media contexts, in order to better understand the effectiveness of hate speech moderation. We hope that our dataset and experimental results will stimulate work in both of these directions to help solve this important problem.

\section*{Limitations}

\paragraph{Dataset}

\textit{Low number of positives}:
The random samples in \textsc{HateDay} used to evaluate hate speech detection in real-world settings contains a low number of hateful examples, ranging from 31 for Kenya to 430 for India. This low number is linked to the very low prevalence of hate speech in our dataset as well as our budget constraint, which impeded us from further expanding the annotation effort. 
While statistically significant, we acknowledge that our performance results on \textsc{HateDay} are necessarily more uncertain, as illustrated by the larger confidence intervals in Tables \ref{tab:os_model_perf_lang} and \ref{tab:os_model_perf_country}.

\textit{Limited generalizability to other platforms, timeframes, and linguistic domains}:
The entirety of our dataset was sampled from one single social media platform for a very bounded timeframe, namely 24 hours. This limits the generalizability of our performance results to data from other social media platforms and covering other timespans.

\textit{Online hate speech is multimodal}: Our work focuses only on text-based hate to limit annotation costs, but we acknowledge that online hate speech is multimodal and that a non-trivial share of this phenomenon, which our analysis necessarily misses, is expressed through images, sounds, or videos \cite{botelho-etal-2021-deciphering,hee2024recent}.

\textit{Limits to representativeness}: The tools used to stratify the \textsc{TwitterDay} dataset into language or country-specific sets, namely language detection and user location inference, are imperfect \cite{hecht2011tweets, graham2014world, jurgens-etal-2017-incorporating}. This implies that the representativeness of \textsc{HateDay} may not be perfect, with language or country sets containing some posts in other languages or from other countries. We acknowledge this limitation, but argue that \textsc{HateDay} remains the most representative samples of Twitter possible with current stratification tools.

\textit{Moderation prior to data collection}: Our analysis assumes that the hateful content in \textsc{HateDay} is representative of all hateful content posted on Twitter on the day of analysis. However, we recognize that some instances of hate speech may have been moderated by Twitter before the data was collected, making our estimates a lower bound. Nonetheless, since the posts were collected 10 minutes after they were posted \cite{pfeffer2023just}, we believe that the enforcement of moderation in such a short timeframe is likely to be minimal.

\paragraph{Experiments}

\textit{Other prompts could lead to different results}:
We craft a prompt using the terms ``hateful'' or ``non-hateful'' (see \ref{app:zero-shot} for details), which exhibit good performance in past research for hate speech detection using zero-shot learning \cite{plaza-del-arco-etal-2023-respectful}. We do not test other prompts and acknowledge that using other prompts may have an impact on classification performance.

\paragraph{Moderation analysis}
We acknowledge that our analysis on the feasibility of moderation is limited in the sense that we operate with publicly available resources while platforms also rely on private models and data which may improve moderation performance for a given cost.

\textit{Private models} While we evaluate publicly available detection models, we are aware that platforms have developed their own detection models, which are kept private and may perform better.

\textit{Private data} Platforms also rely on several data sources to develop their moderation models, such as the user history, user-graph, conversational context, which we do not have access to.  While we consider only public hate speech detection algorithms as a way to flag hate speech in the flow of social media content, we are also aware that platforms use other flagging mechanisms based on private data, such as user reporting or banned word lists, which may increase the share of all hate moderated for a given annotation cost. Finally, the metric platforms aim to reduce is the view count of harmful content such as hate speech, which is arguably more relevant than the sheer prevalence of it from a harm perspective. Unfortunately, such view data is unavailable to us for the considered timeframe.

% \todo{[reformulate: "the small pool of annotators was driven by the logistical constraints of hiring and training them to the required standard and protecting their welfare given the sensitivity and complexity of the topic. Nonetheless, it raises the potential for bias. We take steps to address this in our test set construction and provide an annotator ID with each entry in our publicly-released dataset to enable further research into this issue.]}

% \paragraph{Ambiguity}: \todo{in our annotation process, we found that while politically-based hate adopted the codes of regular hate in some cases (e.g., dehumanization by comparing party followers to animals (cebong in Indonesian, DemocRATS, etc..)), there were also borderline cases between hate and political criticism (saying that a party ruined the country and calling its followers idiots, one is hate, the other not really but can happen in the same sentence).}

\section*{Ethical Considerations}

\paragraph{Annotator wellbeing}

Annotators were provided with clear information regarding the nature of the annotation task before they began their work. They were made aware that they could stop the task at any time if necessary.
% They received a compensation ranging between \todo{X} and \todo{Y} depending on country of residence and professional experience.

\paragraph{Data privacy}

To protect the identity of hateful users and their victims, we anonymize all tweets in our dataset upon release, replacing all usernames by a fixed token @USER. 

\paragraph{Intended use}
The intended use of the \textsc{HateDay} dataset is for research purposes only.

\section*{Acknowledgements}

We thank the annotators for their excellent research assistance.\ 
The study was supported by funding from the United Kingdom’s Foreign Commonwealth and  Development Office (FCDO), the World Bank’s Research Support Budget, and the Gates Foundation (INV057844). This work was also supported in part through the NYU IT High Performance Computing resources, services, and staff expertise. The findings, interpretations, and conclusions expressed in this article are entirely those of the authors. They do not necessarily represent the views of the International Bank for Reconstruction and Development/World Bank and its affiliated organizations or those of the Executive Directors of the World Bank or the governments they represent.  Scott Hale was supported by the ESRC Digital Good Network (grant reference ES/X502352/1). Paul Röttger is a member of the Data and Marketing Insights research unit of the Bocconi Institute for Data Science and Analysis, and is supported by a MUR FARE 2020 initiative under grant agreement Prot. R20YSMBZ8S (INDOMITA).
\bibliography{anthology,custom}
\bibliographystyle{acl_natbib}
\clearpage

\appendix

\section{Twitter Day}\label{app:twitterday}

\subsection{Language distribution}\label{app:twitterday_lang_distribution}

We provide the language share of all posts in \textsc{TwitterDay} \cite{pfeffer2023just} in Table \ref{tab:language_share_twitterday}. We retain the languages for which there was one at least one academic hate speech dataset on \url{https://hatespeechdata.com/} at the time of the analysis (August 2024).

\begin{table}[ht]
    \centering
    \begin{tabular}{@{}lc@{}}
        \toprule
        \textbf{Language} & \textbf{Share (\%)} \\ \midrule
        \textbf{English}  & 27.6  \\
        Japanese  & 20.9  \\
        \textbf{Spanish}  &  6.7  \\
        \textbf{Arabic}  &  6.6  \\
        \textbf{Portuguese}  &  5.4  \\
        \textbf{Indonesian}  &  2.9  \\
        Korean  &  2.5  \\
        \textbf{Turkish}  &  2.2  \\
        Farsi  &  1.6  \\
        \textbf{French}  &  1.6  \\
        Thai  &  1.2  \\
        Tagalog  &  0.9  \\
        \textbf{German}  &  0.7  \\ \bottomrule
    \end{tabular}
    \caption{Share of all original tweets (dropping retweets) in \textsc{TwitterDay} by language. Retained languages are bolded.}
    \label{tab:language_share_twitterday}
\end{table}
%%%%%%%%%%%%%%%%%%%%%%%%%%%%%%%%%%%%%%%%%%%%%%%%%%%%%%%%%%%%%%%%%%%%%%%%%%%%%%%%%%%%%%
%%%%%%%%%%%%%%%%%%%%%%%%%%%%%%%%%%%%%%%%%%%%%%%%%%%%%%%%%%%%%%%%%%%%%%%%%%%%%%%%%%%%%%
\section{Annotation}\label{app: annotation}

% Following \citet{bender-friedman-2018-data}, we provide a data statement, which documents the provenance and annotation of the \textsc{HateDay} dataset.

\subsection{Annotator demographics}\label{app: annotator demographics}
We recruit a team of 30 annotators, that is three for each  language and country. We provide detailed demographics for each language and country below:

\paragraph{Arabic}
We recruit three Arabic-speaking male annotators. All three annotators are 18-29. Two annotators are educated to undergraduate level and the other one to taught masters. All three are native Arabic speakers respectively from Egypt, Tunisia and the United Arab Emirates.

\paragraph{English}
We recruit two English-speaking female annotators and one male. Two annotators out of three are 18-29 while the other one is 30-39. One annotator is educated to undergraduate level, another one to taught masters and the last one to research degree~(i.e. PhD). Two are native English speakers, respectively from the United States and Nigeria, and one is a non-native German citizen fluent in English.

\paragraph{French}
We recruit two French-speaking male annotators and one female. One annotator out of three is 18-29 while the other two are 30-39. All annotators are educated to taught masters level. All three are native French speakers. Two annotators are French citizens while the last one is Canadian.

\paragraph{German}
We recruit three German-speaking male annotators. Two annotators out of three are 18-29 while the other one is 30-39. Two annotators are educated to undergraduate level and the last one to research degree~(i.e. PhD). All three are native German speakers, all from Germany.

\paragraph{Indonesian}
We recruit two Indonesian-speaking male annotators and one female. Two annotators out of three are 18-29 while the other one is 30-39. Two annotators are educated to undergraduate level and the last one to research degree~(i.e. PhD). All three are native Indonesian speakers, all from Indonesia.

\paragraph{Portuguese}
We recruit two Portuguese-speaking female annotators and one male. All annotators are 30-39 and are educated to taught masters level. Two are native Portuguese speakers from Brazil and one is a non-native Portuguese-speaking Mexican citizen fluent in Portuguese from Portugal.

\paragraph{Spanish}
We recruit two Spanish-speaking female annotators and one male. All annotators are 30-39. Two annotators are educated to undergraduate level and the last one to taught masters. All three are native Spanish speakers from Mexico.

\paragraph{Turkish}
We recruit two Turkish-speaking male annotators and one female. Two annotators are 18-29 and the last one is 30-39. One annotator is educated to undergraduate level and the other two to taught masters. All three are native Turkish speakers from Turkey.

\paragraph{United States}
We recruit two female annotators and one male. Two annotators out of three are 18-29 while the other one is 30-39. Two annotators are educated to taught masters and the last one to research degree~(i.e. PhD). Two annotators are American citizens while the last one is a German citizen residing in the United States for more than five years.

\paragraph{India}
We recruit two male annotators and one female. All three annotators are 18-29 and are educated to undergraduate level. All three annotators are Indian citizens.

\paragraph{Nigeria}
We recruit two female annotators and one male. All three annotators are 30-39 and are educated to undergraduate level. All three annotators are Nigerian citizens.

\paragraph{Kenya}
We recruit two female annotators and one male. All three annotators are 30-39 and are educated to undergraduate level. All three annotators are Kenyan citizens.

% All annotators had previously completed annotation work on at least one other hate speech dataset.
% In the first week, we introduced the binary annotation task to them in an onboarding session and tested their understanding on a set of 100 cases, which we then provided individual feedback on.
% In the second week, we asked each annotator to annotate around 2,000 test cases so that each case in our test suite was annotated by varied sets of exactly five annotators.
% Throughout the process, we communicated with annotators in real-time over a messaging platform.
% We also followed guidance for protecting and monitoring annotator well-being provided by \citet{vidgen2019challenges}.

\subsection{Annotation guidelines}\label{app:annotation_guidelines}
We provide a version of the annotation guidelines at\ \url{https://github.com/manueltonneau/hateday_acl/blob/main/annotation_guidelines.pdf}

\subsection{Recruitment and compensation} Annotators were recruited through email outreach. Net hourly rate ranged from 5 to 24 USD depending on country of residence, education and experience.

% \subsection{Agreement rates}\label{app:agreement_rates}

\section{Academic Hate Speech Datasets}

\subsection{Language-level Supersets}\label{app:language_supersets}

The language-level supersets are documented in \citet{tonneau-etal-2024-languages} and obtained from Hugging Face.\footnote{\url{https://huggingface.co/collections/manueltonneau/hate-speech-supersets-664ef6d2bc40cce7a8b1092f}} The number of annotated datapoints in these supersets for each language is:
\begin{itemize}
    \item Arabic: 464,260
    \item English: 590,142
    \item French: 18,071
    \item German: 60,680
    \item Indonesian: 14,306
    \item Portuguese: 43,222
    \item Spanish: 34,811
    \item Turkish: 115,408
\end{itemize}

\subsection{Country-level Dataset Survey}\label{app:data_survey}

In the absence of supersets at the country-level, we conduct a survey of hate speech datasets grounded geographically in India, Nigeria and Kenya. 

\subsubsection{Surveying approach}
To identify HS datasets, we rely on three data sources. First, we inspect the Hate Speech Data Catalogue\footnote{\url{https://hatespeechdata.com/}} \cite{vidgen2020directions}.
Second, we inspect the datasets listed in the latest survey of hate speech datasets \cite{poletto2021resources}. 
Finally, we conduct a Google search for each country and inspect the links of the first three result pages in each case. We keep only the datasets that fit the following three criteria: 
 \begin{enumerate}
     \vspace*{-0.5em}
  \itemsep-0em 
     \item The dataset is documented, meaning it is attached to a research paper or a README file describing its construction.
     \item The dataset is either publicly available or could be retrieved after contacting the authors.
     \item The dataset focuses on hate speech, defined broadly as ``any kind of communication in speech, writing or behavior, that attacks or uses pejorative or discriminatory language with reference to a person or a group on the basis of who they are, in other words, based on their religion, ethnicity, nationality, race, color, descent, gender or other identity factor'' \cite{un2019}.
    \vspace*{-0.5em}
\end{enumerate}

\subsubsection{Target categorization}\label{app:target_categorization}

We categorize each surveyed dataset in terms of the target categories they focus on. To do so, we look for mentions of target focus in the data documentation. If it is absent, we inspect the sampling approach and specifically the keywords used for sampling and use these to determine the target focus.

\subsubsection{Surveyed datasets}

We list below the retained datasets for each country, as well as their target focus:

\paragraph{India}
\begin{enumerate}

    \item \textit{A Dataset of Hindi-English Code-Mixed Social Media Text for Hate Speech Detection} \cite{bohra-etal-2018-dataset}. Target focus: race/ethnicity/national origin, gender, caste, religion
    \item \textit{Overview of the HASOC track at FIRE 2019: Hate Speech and Offensive Content Identification in Indo-European Languages} \cite{mandl2019overview}. Target focus: religion, caste, gender, politics
    \item \textit{Hostility Detection Dataset in Hindi} \cite{bhardwaj2020hostility}. Target focus: race/ethnicity/national origin, religion
    \item \textit{Listening to Affected Communities to Define Extreme Speech: Dataset and Experiments} \cite{maronikolakis-etal-2022-listening}. Target focus: religion, caste
    \item \textit{Uncovering Political Hate Speech During Indian Election Campaign: A New Low-Resource Dataset and Baselines} \cite{jafri2023uncovering}. Target focus: politics

\end{enumerate}

We identify two additional datasets \cite{mathur-etal-2018-offend,saroj-pal-2020-indian} that we leave out of the analysis as we could not retrieve them.

\paragraph{Nigeria}
\begin{enumerate}
    \item \textit{HERDPhobia: A Dataset for Hate Speech against Fulani in Nigeria} \cite{aliyu2022herdphobia}. Target focus: race/ethnicity/national origin
    \item \textit{Detection of Hate Speech Code Mix Involving English and Other Nigerian Languages} \cite{ndabula2023detection}. Target focus: race/ethnicity/national origin, religion, politics
    \item \textit{EkoHate: Abusive Language and Hate Speech Detection for Code-switched Political Discussions on Nigerian Twitter} \cite{ilevbare-etal-2024-ekohate}. Target focus: politics
    \item \textit{NaijaHate: Evaluating Hate Speech Detection on Nigerian Twitter Using Representative Data} \cite{tonneau-etal-2024-naijahate}. Target focus: race/ethnicity/national origin, gender, sexual orientation, religion
\end{enumerate}

\paragraph{Kenya}
\begin{enumerate}
    \item \textit{Building and annotating a codeswitched hate speech corpora} \cite{ombui2021building}. Target focus: race/ethnicity/national origin, politics
    \item \textit{Listening to Affected Communities to Define Extreme Speech: Dataset and Experiments} \cite{maronikolakis-etal-2022-listening}. Target focus: race/ethnicity/national origin, religion, politics
\end{enumerate}

%%%%%%%%%%%%%%%%%%%%%%%%%%%%%%%%%%%%%%%%%%%%%%%%%%%%%%%%%%%%%%%%%%%%%%%%%%%%%%%%%%%%%%
%%%%%%%%%%%%%%%%%%%%%%%%%%%%%%%%%%%%%%%%%%%%%%%%%%%%%%%%%%%%%%%%%%%%%%%%%%%%%%%%%%%%%%
\section{Models}

\subsection{Hugging Face models}\label{app:hf-models}

We list the Hugging Face models used as benchmark below for each language and country:

\paragraph{Arabic}
\begin{itemize}
    \item {\tt IbrahimAmin-marbertv2-finetuned-\\egyptian-hate-speech-detection } \cite{10009167}
    \item {\tt Hate-speech-CNERG/dehatebert-mono-\\arabic} \cite{aluru2020deep}

\end{itemize}

\paragraph{English}
\begin{itemize}
    \item {\tt Hate-speech-CNERG/bert-base-uncased-\\hatexplain } \cite{mathew2021hatexplain}
    \item {\tt facebook/roberta-hate-speech-\\dynabench-r4-target} \cite{vidgen-etal-2021-learning}
    \item {\tt Hate-speech-CNERG/dehatebert-mono-\\english} \cite{aluru2020deep}
    \item {\tt IMSyPP/hate\_speech\_en} \cite{kralj2022handling}
    \item {\tt pysentimiento/bertweet-hate-speech} \cite{perez2021pysentimiento}
\end{itemize}

\paragraph{French}
\begin{itemize}
    \item {\tt Hate-speech-CNERG/dehatebert-mono-\\french} \cite{aluru2020deep}
    \item {\tt Poulpidot/distilcamenbert-french-\\hate-speech}
    \item {\tt julio2027/French\_hate\_speech\_\\CamemBERT\_v3}
\end{itemize}

\paragraph{German}
\begin{itemize}
    \item {\tt jagoldz/gahd} \cite{goldzycher-etal-2024-improving}
    \item {\tt deepset/bert-base-german-\\cased-hatespeech-GermEval18Coarse}
    \item {\tt jorgeortizv/BERT-hateSpeech\\Recognition-German}
    \item {\tt Hate-speech-CNERG/dehatebert-\\mono-german} \cite{aluru2020deep}
    \item {\tt shahrukhx01/gbert-hasoc-german-2019}
\end{itemize}

\paragraph{Indonesian}
\begin{itemize}
    \item {\tt Hate-speech-CNERG/dehatebert-mono-\\indonesian } \cite{aluru2020deep}
\end{itemize}

\paragraph{Portuguese}
\begin{itemize}
    \item {\tt knowhate/HateBERTimbau-yt-tt} %\todo{[citation to appear soon on HF]}
    \item {\tt Hate-speech-CNERG/dehatebert-mono-\\portugese} \cite{aluru2020deep}
\end{itemize}

\paragraph{Spanish}
\begin{itemize}
    \item {\tt Hate-speech-CNERG/dehatebert-mono\\-spanish} \cite{aluru2020deep}
    \item {\tt pysentimiento/robertuito-hate-speech} \cite{perez2021pysentimiento}
    \item {\tt jorgeortizfuentes/chilean-spanish-\\hate-speech}
\end{itemize}

\paragraph{Turkish}
\begin{itemize}
    \item {\tt ctoraman/hate-speech-berturk} \cite{toraman-etal-2022-large}
\end{itemize}

\paragraph{United States}
We use the same models as for English for the United States.

\paragraph{India}
\begin{itemize}
    \item {\tt Hate-speech-CNERG/kannada-codemixed-\\abusive-MuRIL} \cite{das2022data}
    \item {\tt Hate-speech-CNERG/marathi-codemixed-\\abusive-MuRIL} \cite{das2022data}
    \item {\tt Hate-speech-CNERG/bengali-abusive-\\MuRIL} \cite{das2022data}
    \item {\tt Hate-speech-CNERG/tamil-codemixed-\\abusive-MuRIL} \cite{das2022data}
    \item {\tt Hate-speech-CNERG/english-abusive-\\MuRIL} \cite{das2022data}

\end{itemize}

\paragraph{Nigeria}
\begin{itemize}
    \item {\tt worldbank/naija-xlm-twitter-base-hate} \cite{tonneau-etal-2024-naijahate}
\end{itemize}

\paragraph{Kenya}
We have not found any hate speech detection models tailored to the Kenyan context on Hugging Face.

\subsection{Zero-shot learning}\label{app:zero-shot}

The prompt we use for zero-shot learning: ``\textit{Here is a social media post: [POST]. Classify this post as ``hateful'' or ``non-hateful''. Only return the predicted class.}''.

The prompt is run three times for each tweet. We then define the hateful score as the share of the three times for which the model predicted that the tweet was hateful. We then use this score to compute the average precision. We use all default values for the main hyperparameters, including 1 for temperature.

% \subsection{Model evaluation}

% \todo{add details on training and eval}

\subsection{Computing infrastructure}

For inference, we used either V100 (32GB) or RTX8000 (48GB) GPUs.  

\subsection{Number of parameters}
The supervised learning models are largely based on the BERT base architecture which has 110 million parameters. The decoder-based models (Aya and Llama3.1) both have 8 billion parameters.

%%%%%%%%%%%%%%%%%%%%%%%%%%%%%%%%%%%%%%%%%%%%%%%%%%%%%%%%%%%%%%%%%%%%%%%%%%%%%%%%%%%%%%
%%%%%%%%%%%%%%%%%%%%%%%%%%%%%%%%%%%%%%%%%%%%%%%%%%%%%%%%%%%%%%%%%%%%%%%%%%%%%%%%%%%%%%
\section{Additional Results}

\subsection{Figures}
\begin{figure*}[ht]
    \raggedleft
    \includegraphics[width=\textwidth]{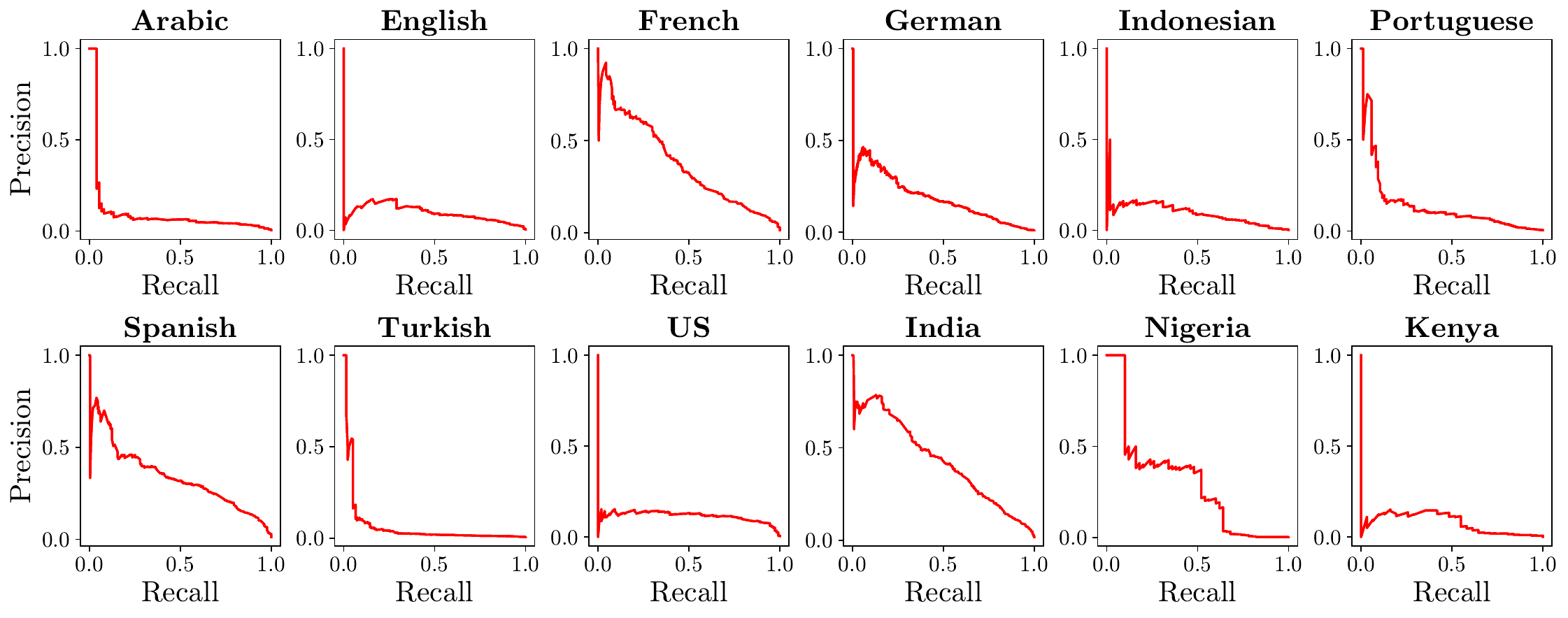} 
    \caption{Precision-recall curves for each language and country}
    \label{fig:precision_recall}
\end{figure*}

\begin{figure*}[ht]
    \raggedleft
    \includegraphics[width=1\textwidth]{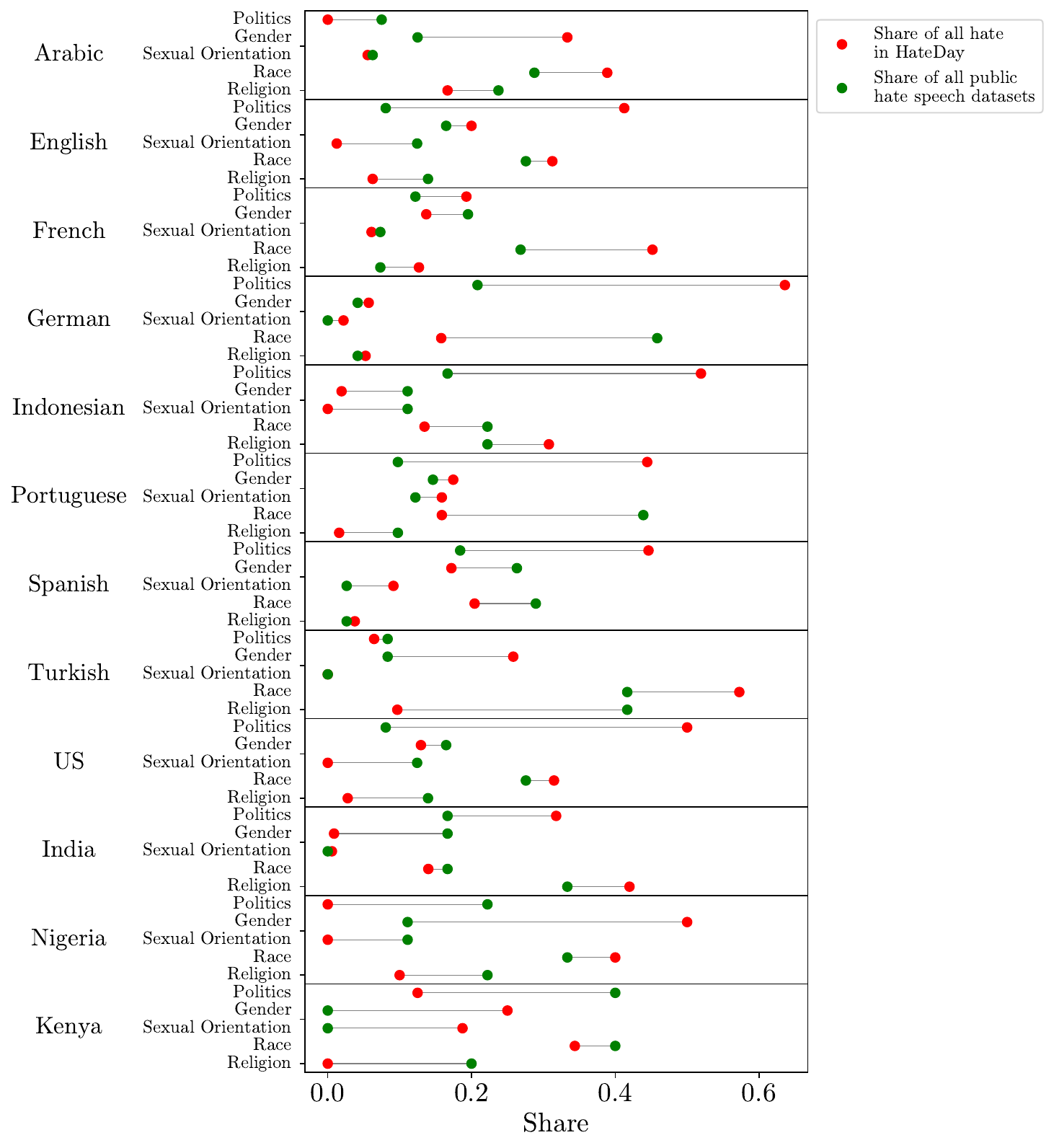} 
    \caption{Comparison between the target-level share of all hate in \textsc{HateDay} and the share of all hate speech datasets for each language or country and target combination. The language-level data for hate speech datasets is taken from \citet{yu2024unseen}. }
    \label{fig:comparison_prevalence_academic}
\end{figure*}

\begin{figure*}[ht]
    %\raggedleft
    \centering
    \includegraphics[width=0.75\textwidth]{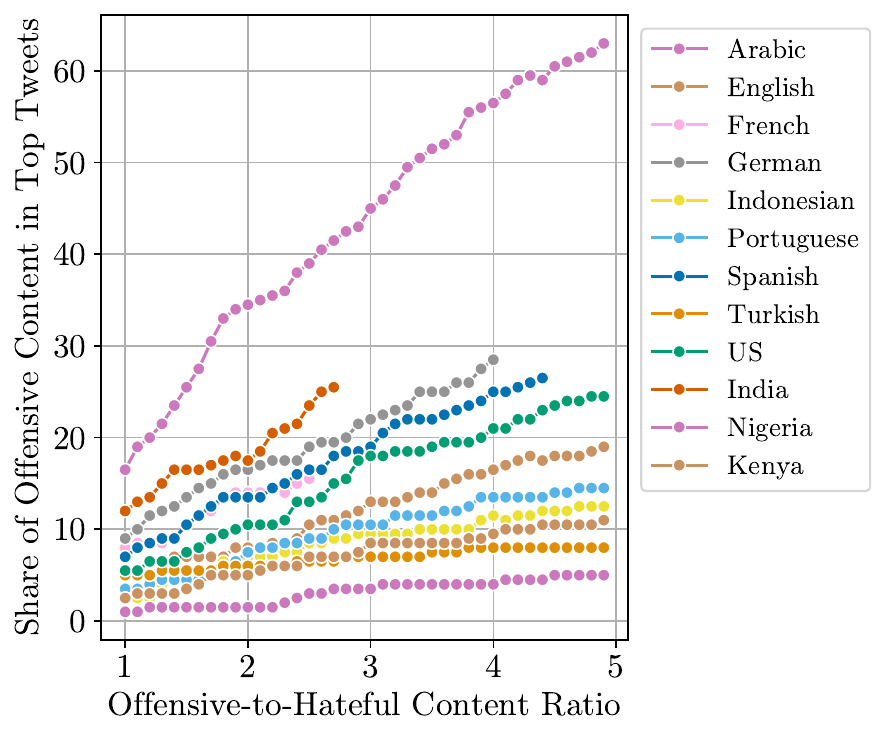} 
    \caption{Offensive-to-hateful count ratio versus the share of offensive content in top tweets (in \%). Top tweets are defined as the 1\% (N=200) top-scored tweets in \textsc{HateDay} using the best performing model on the dataset.}
    \label{fig:increasing_offensive_ratio}
\end{figure*}

We provide additional figures representing respectively the precision-recall curves across languages and countries (Figure \ref{fig:precision_recall}), the comparison between target-level academic focus and target-level share of all hate in \textsc{HateDay} (Figure \ref{fig:comparison_prevalence_academic}) and the relationship between the offensive-to-hateful count ratio and the share of offensive content in top-scored tweets of \textsc{HateDay} (Figure \ref{fig:increasing_offensive_ratio}). 

\subsection{Individual open-source model performance}\label{app:indiv_model_perf} We provide detailed average precision results on the three evaluation sets for each individual open-source model (see \ref{app:hf-models} for the full list) and across languages (Table \ref{tab:os_model_perf_lang}) and countries (Table \ref{tab:os_model_perf_country}).

\begin{table*}[ht]
    \centering
    \begin{tabular}{l l >{\columncolor[rgb]{0.9, 0.95, 1}}c >{\columncolor[rgb]{0.9, 1, 0.9}}c >{\columncolor[rgb]{0.95, 0.9, 1}}c}
        \hline
        \textbf{Language} & \textbf{Model} &  \textbf{AD} & \textbf{HC} & \textbf{HD} \\
        \hline
        Arabic & \textit{IbrahimAmin/marbertv2-finetuned-} & & &\\
         & \textit{egyptian-hate-speech-detection} & \textbf{25.8±0.6} & 76.9 \newline $\pm$2.0 & 5.8 \newline $\pm$3.0 \\
         & \textit{Hate-speech-CNERG- \newline dehatebert-mono-arabic} & 16.4±0.4 & 83.1$\pm$1.7 & 1.6$\pm$1.4 \\
          & \textit{Perspective API} & 18.7±0.4 & \textbf{89.6$\pm$1.1} & \textbf{10.2$\pm$5.6} \\
        \hline
        
        English & \textit{Hate-speech-CNERG/bert-base-uncased-hatexplain} & 47.2±0.3 & 74.3±2.0 & 2.5±1.0 \\
         & \textit{facebook/roberta-hate-speech-dynabench-r4-target} & \textbf{55.3±0.3} & \textbf{98.7±0.4} & 4.2±2.2 \\
          & \textit{Hate-speech-CNERG/dehatebert-mono-english} & 48.7±0.3 & 75.0±2.0 & 7.2±4.4 \\
          & \textit{IMSyPP/hate\_speech\_en} & 21.7±0.1 & 65.9±2.0 & 0.6±0.2 \\
          & \textit{pysentimiento/bertweet-hate-speech} & 38.2±0.3 & 82.2±1.7 & 9.0±3.9 \\
          & \textit{Perspective API} & 52.9±0.3 & 95.1±0.6 & \textbf{10.1±3.5} \\
        \hline
        
        French & \textit{Hate-speech-CNERG/dehatebert-mono-french} & 27.5±1.0 & 74.5±1.9 & 2.7±0.6 \\
         & \textit{Poulpidot/distilcamenbert-french-hate-speech} & 27.6±0.9 & 87.2±1.5 & 4.2±0.9 \\
          & \textit{julio2027/French\_hate\_speech\_CamemBERT\_v3} & 35.7±1.3 & 89.9±1.2 & 16.0±4.2\\
          & \textit{Perspective API} & \textbf{40.1±1.4} & \textbf{96.9±0.4} & \textbf{37.7±6.0}  \\
        \hline
        
        German & \textit{jagoldz-gahd} & \textbf{64.3±1.2} & \textbf{97.2±0.5} & \textbf{19.3±5.0} \\
         & \textit{deepset/bert-base-german-cased-hatespeech-}&  &  &  \\
        & \textit{GermEval18Coarse} & 22.6±0.8 & 82.1±1.9 & 8.5±2.6 \\
          & \textit{jorgeortizv/BERT-hateSpeechRecognition-German} & 19.9±0.7 & 76.8±2.0 & 3.4±1.1 \\
          & \textit{Hate-speech-CNERG/dehatebert-mono-german} & 18.1±0.7 & 80.2±1.9 & 1.8±0.5 \\
        & \textit{shahrukhx01-gbert-hasoc-german-2019} & 19.0±0.7 & 79.4±1.8 & 5.6±1.4 \\
          & \textit{Perspective API} & 53.6±1.1 & 96.0±0.5 & 18.8±4.4\\
        \hline
        
        Indonesian & \textit{Hate-speech-CNERG-dehatebert-mono-indonesian
} & \textbf{90.2±0.7} & - & 3.5±2.3 \\
          & \textit{Perspective API} & 65.5±1.3 & - & \textbf{11.1±5.9} \\
        \hline
        Portuguese & \textit{knowhate/HateBERTimbau-yt-tt} & 18.7±0.9 & 85.8±1.6 & 3.1±1.7 \\
         & \textit{Hate-speech-CNERG/dehatebert-mono-portugese} & 19.1±1.0 & 74.4±1.8 & 1.3±0.5\\
          & \textit{Perspective API} & \textbf{30.9±1.5} & \textbf{94.6±0.7} & \textbf{14.5±6.9} \\
        \hline
        
        Spanish & \textit{Hate-speech-CNERG/dehatebert-mono-spanish} & 60.3±1.0 & 78.2±1.9 & 3.3±1.0 \\
         & \textit{pysentimiento/robertuito-hate-speech} & \textbf{72.6±0.9} & 87.0±1.5 & 8.2±1.6 \\
         & \textit{jorgeortizfuentes/chilean-spanish-hate-speech} & 54.9±1.1 & 82.3±1.6 & 11.6±3.3 \\
          & \textit{Perspective API} & 50.9±1.1 & \textbf{96.0±0.5} & \textbf{34.1±6.2}  \\
        \hline
        
        Turkish & \textit{ctoraman/hate-speech-berturk} & \textbf{32.1±0.4} & - & 1.9±0.6 \\
          & \textit{Perspective API} & 31.4±0.4 & - & \textbf{6.5±3.8} \\
        \hline
    \end{tabular}
    \caption{\textbf{Detailed open-source and Perspective model performance across languages and evaluation sets}, as measured by average precision (\%). Metrics are reported with 95\% bootstrapped confidence intervals. We report performance on three evaluation sets: \colorbox{holdout}{academic hate speech datasets (AD)} combined for a given language, \colorbox{hatecheck}{HateCheck functional tests (HC)} and \colorbox{hateday}{\textsc{HateDay}} \colorbox{hateday}{(HD)}. HC does not cover Indonesian and Turkish.}
    \label{tab:os_model_perf_lang}
\end{table*}

\begin{table*}[ht]
    \centering
    \begin{tabular}{l l >{\columncolor[rgb]{0.9, 0.95, 1}}c >{\columncolor[rgb]{0.95, 0.9, 1}}c}
        \hline
        \textbf{Country} & \textbf{Model} &  \textbf{AD} & \textbf{HD} \\
        \hline
        
        United States & \textit{Hate-speech-CNERG/bert-base-uncased-hatexplain} & 47.2±0.3 & 4.6±2.3 \\
         & \textit{facebook/roberta-hate-speech-dynabench-r4-target} & \textbf{55.3±0.3} & 3.1±0.9  \\
          & \textit{Hate-speech-CNERG/dehatebert-mono-english} & 48.7±0.3 & 4.6±2.3 \\
          & \textit{IMSyPP/hate\_speech\_en} & 21.7±0.1 &  0.6±0.2 \\
          & \textit{pysentimiento/bertweet-hate-speech} & 38.2±0.3 & 8.2±2.7 \\
          & \textit{Perspective API} & 52.9±0.3 & \textbf{12.3±3.4} \\
        \hline
        
        India & \textit{Hate-speech-CNERG/kannada-codemixed-abusive-MuRIL} & 48.6±1.1 & 5.9±1.1 \\
         & \textit{Hate-speech-CNERG/marathi-codemixed-abusive-MuRIL} & 43.4±1.0 & 3.7±0.9 \\
          & \textit{Hate-speech-CNERG/bengali-abusive-MuRIL} & 40.0±1.0 & 3.3±0.7 \\
          & \textit{Hate-speech-CNERG/tamil-codemixed-abusive-MuRIL} & 47.1±1.0 & 6.2±1.2\\
          & \textit{Hate-speech-CNERG/english-abusive-MuRIL} & 54.3±1.1 & 7.8±1.6 \\
          & \textit{Perspective API} & \textbf{61.9±1.1} &  \textbf{42.9±5.8} \\
        \hline
        
        Nigeria & \textit{worldbank/naija-xlm-twitter-base-hate} & \textbf{65.7±1.4} & \textbf{30.9±1.3}  \\
          & \textit{Perspective API} & 44.1±1.6 & 8.6±6.5 \\
        \hline
        Kenya  & \textit{Perspective API} & 31.6±0.9 & 9.1±6.1 \\
        \hline
    \end{tabular}
    \caption{\textbf{Detailed open-source and Perspective model performance across countries and evaluation sets}, as measured by average precision (\%). Metrics are reported with 95\% bootstrapped confidence intervals. We report performance on two evaluation sets: \colorbox{holdout}{academic hate speech datasets (AD)} combined for a given language and \colorbox{hateday}{\textsc{HateDay}} \colorbox{hateday}{(HD)}.}
    \label{tab:os_model_perf_country}
\end{table*}

\subsection{Human-in-the-loop moderation}\label{app:human-in-the-loop}

In the human-in-the-loop setting, we compute the proportion of all tweets to be reviewed by human annotators as follows. We first compute the number of predicted positives PP:
$$PP = TP + FP = TP/\textrm{precision}$$
$$PP = (\textrm{recall} * (TP+FN))/\textrm{precision}$$
$$PP = (\textrm{recall} * \textrm{total \# hateful tweets})/\textrm{precision}$$
$$PP = (\textrm{recall} * \textrm{base rate} * \textrm{total \# tweets})/\textrm{precision}$$

We then derive the share $S$ of all tweets that are predicted positive by a given model, that is the share of all tweets that will be reviewed by human moderators in a human-in-the-loop approach, by dividing $PP$ by the total number of tweets:

$$S = (\textrm{recall} * \textrm{base rate})/\textrm{precision}$$

\noindent with the base rate equal to the prevalence of hateful content in \textsc{HateDay} for that specific language. We finally use the precision and recall values from the precision-recall curve to derive the curve illustrating the relationship between recall and $S$. 

\end{document}